\documentclass[10pt,twocolumn,letterpaper]{article}

\usepackage{amsmath}
\usepackage{amssymb}
\usepackage{wasysym}
\usepackage{amssymb}
\usepackage{amsthm}
\usepackage{color}
\usepackage{amsmath}
\usepackage{amsmath}
\usepackage{amssymb}
\usepackage{amsthm}
\usepackage{amsmath}
\usepackage{comment}
\usepackage{graphicx}
\usepackage{float}
\usepackage{multirow}
\usepackage{epsfig}
\usepackage{graphicx}
\usepackage[english]{babel}
\usepackage{cvpr}
\usepackage{times}
\usepackage{epsfig}
\usepackage{amsmath}
\usepackage{amssymb}
\usepackage{enumerate}
\usepackage[titletoc]{appendix}

\newtheorem{theorem}{Theorem}[]

\newtheorem{lemma}[theorem]{Lemma} 

\newtheorem{definition}{Definition}[]

\newcommand{\Real}{\mathbb R}
\newcommand{\Symmetric}{\mathbb S}
\newcommand{\rank}{\text {rank}}
\newcommand{\diag}{\text {diag}}
\newcommand{\norm}[1]{\left\lVert #1 \right\rVert}
\newcommand{\av}{\mathbf{a}}
\newcommand{\bv}{\mathbf{b}}
\newcommand{\xx}{\mathbf{x}}
\newcommand{\yy}{\mathbf{y}}
\newcommand{\tr}{\mathbf{t}}
\newcommand{\ep}{\mathbf{e}}
\newcommand{\uu}{\mathbf{u}}
\newcommand{\vv}{\mathbf{v}}

\definecolor{mypink1}{rgb}{0.33, 0.76, 0.66}
\newcommand{\yk}[1]{\textcolor{mypink1}{[Yoni: #1]}}
\newcommand{\rb}[1]{\textcolor{blue}{[Ronen: #1]}}

\newcommand{\ag}[1]{\textcolor{red}{[Amnon: #1]}}

\usepackage[pagebackref=true,breaklinks=true,letterpaper=true,colorlinks,bookmarks=false]{hyperref}

\cvprfinalcopy % *** Uncomment this line for the final submission

\def\cvprPaperID{8232} % *** Enter the CVPR Paper ID here

\ifcvprfinal\fi
\begin{document}

%%%%%%%%% TITLE
\title{Averaging Essential and Fundamental Matrices in Collinear Camera Settings}
\author{Amnon Geifman* \hspace{1cm}Yoni Kasten* \hspace{1cm} Meirav Galun  \hspace{1cm}Ronen Basri \\
Weizmann Institute of Science\\
{\tt\small  \{amnon.geifman,yoni.kasten,meirav.galun,ronen.basri\}@weizmann.ac.il}
}
\maketitle
\makeatletter
\def\blfootnote{\xdef\@thefnmark{}\@footnotetext}
\makeatother
\blfootnote{*Equal contributors}
%\thispagestyle{empty}

%%%%%%%%% ABSTRACT
\begin{abstract}
Global methods to Structure from Motion have gained popularity in recent years. A significant drawback of global methods is their sensitivity to collinear camera settings. In this paper, we introduce an analysis and algorithms for averaging bifocal tensors (essential or fundamental matrices) when either subsets or all of the camera centers are collinear.
 We provide a complete spectral characterization of bifocal tensors in collinear scenarios and further propose two averaging algorithms. The first algorithm uses rank constrained minimization to recover camera matrices in fully collinear settings. The second algorithm enriches the set of possibly mixed collinear and non-collinear cameras with additional, ``virtual cameras," which are placed in general position, enabling the application of existing averaging methods to the enriched set of bifocal tensors. Our algorithms are shown to achieve state of the art results on various benchmarks that include autonomous car datasets and unordered image collections  in both calibrated and unclibrated settings.        \end{abstract}

%%%%%%%%% BODY TEXT

%-------------------------------------------------------------------------

\medskip

% \begin{figure}[t]
% \begin{center}
%    %\includegraphics[width=0.8\linewidth]{egfigure.eps}
% \end{center}
%    \caption{Example of caption.  It is set in Roman so that mathematics
%    (always set in Roman: $B \sin A = A \sin B$) may be included without an
%    ugly clash.}
% \label{fig:long}
% \label{fig:onecol}
% \end{figure}
% 

% \begin{figure*}
% \begin{center}
% \fbox{\rule{0pt}{2in} \rule{.9\linewidth}{0pt}}
% \end{center}
%    \caption{Example of a short caption, which should be centered.}
% \label{fig:short}
% \end{figure*}

%------------------------------------------------------------------------
\section{Introduction}

Global approaches to Structure from Motion (SfM) use bifocal tensors (essential or fundamental matrices) between pairs of images to recover camera parameters in multiview settings. These methods have  gained popularity in recent years due to their high accuracy and improved efficiency. In contrast to incremental methods, which recover camera parameters for one image at a time and thus involve repeated application of bundle adjustment (BA) for each handled image, global algorithms apply BA only once, considerably reducing execution time. Existing global algorithms largely proceed in two steps, applying rotation averaging followed by translation averaging. Recent algorithms further improve accuracy by directly averaging essential and fundamental matrices in one step \cite{kasten2019algebraic,kasten2018gpsfm}. 

A significant drawback of global methods is their sensitivity to collinear camera settings. When all camera centers in a scene lie along a line, bifocal tensors do not determine camera locations along this line, and point matches across three or more images must be utilized. Moreover, with only bifocal tensors, subsets of collinear cameras can lead to reconstructions of scene parts that are attached nonrigidly.
Finally, the averaging algorithms of \cite{kasten2019algebraic,kasten2018gpsfm} critically base their recovery on triplet sub-collections of images whose cameras must lie in general position. This severely limits the applicability of these algorithms, requiring in many cases to remove many images from the input datasets. Handling collinear camera settings is critical to many SLAM applications, including autonomous driving  \cite{Geiger2012CVPR}. 

This paper introduces an analysis and novel solutions to 3D reconstruction problems involving cameras with collinear centers in the context of bifocal tensor averaging. We note that to date this problem has been addressed only in the context of translation averaging \cite{cui2015linear,jiang2013global,wilson2014robust}. We introduce a complete algebraic characterization of bifocal tensors  in collinear scenarios, providing both necessary and sufficient conditions that bifocal tensors can be realized by cameras with collinear centers. Our analysis complements the conditions derived for cameras in general position in \cite{kasten2019algebraic,kasten2018gpsfm} and the partial conditions for collinear settings derived in \cite{sengupta2017new}. Specifically,  adopting the definitions of $n$-view bifocal matrices introduced in those papers, we provide a full characterization in terms of spectral decomposition and rank patterns of these matrices.

We build upon this characterization to design state of the art algorithms for global SfM that are applicable in both calibrated and uncalibrated settings. We first introduce a method that, given possibly erroneous bifocal tensors, enforces our spectral constraints. This algorithm is suitable for image collections captured in fully collinear settings.

We subsequently present a second algorithm  for bifocal tensor averaging that can incorporate both collinear cameras and cameras in general position. This algorithm is based on the following novel observation. Given a point match across three views, it is possible to define a \textit{virtual} camera centered at the unknown 3D location of this point and subsequently construct bifocal tensors relating this virtual camera with any of the three cameras corresponding to these views. Choosing this point so that its projections lie away from the epipoles ensures that the center of the virtual camera is non-collinear with the real cameras. We can therefore augment the set of bifocal tensors with the newly constructed matrices and then feed them to standard bifocal tensors averaging schemes, allowing us to obtain solutions in both fully and partly collinear camera settings.
%They  get as input a noisy $n$-view fundamental/ essential matrices and seeks to find the closest consistent $n$-view matrix to the measured matrix. 

%Practically, we also suggest a novel method to cope with collinear  motion by adding points to the viewing graph.  This method is the first to formulate an essential or fundamental matrix w.r.t a 3D point. In the method  formulation, given a set of collinear cameras we define a new view  consists from a 3-D point (which in general is not collinear with the other views), and we insert the corresponding essential/fundamental matrices to the $n$-view matrix . This enables us to use the broad theoretical and practical work done by \cite{kasten2018gpsfm,kasten2019algebraic} and to design an averaging algorithm for the $n$-view matrix based on the non-collinear formulation.     

We demonstrate state of the art results in various applications, and specifically improve over the recent results of \cite{kasten2019algebraic,kasten2018gpsfm} by allowing to incorporate collinear camera triplets in the optimization process. We evaluate our proposed algorithms on four benchmarks: autonomous car datasets \cite{Geiger2012CVPR} and unordered collections of images \cite{wilson2014robust,olsson2011stable,vgg_dataset} in both calibrated and uncalibrated settings.

\section{Related work}

Incremental approaches for calibrated \cite{schonberger2016structure,klopschitz2010robust,snavely2008modeling,agarwal2009building,wu2013towards} and uncalibrated SfM settings \cite{magerand2017practical,pollefeys2004visual} use two images to obtain an initial reconstruction and then incrementally use camera resectioning methods  \cite{gao2003complete,kasten2019resultant,persson2018lambda,zheng2015structure,hartley2003multiple,camposeco2018hybrid}, adding one image at a time to expand the reconstruction. Bundle Adjustment \cite{triggs1999bundle} is performed  with every additional image to prevent error drift of camera parameters, rendering this process computationally demanding.

Global approaches to SfM use collections of bifocal tensors to  simultaneously solve for the parameters of all cameras, and subsequently perform a single round of  Bundle Adjustment. Most existing global approaches for calibrated settings first extract the pairwise rotations from the essential matrices, then perform rotation averaging \cite{arie2012global,martinec2007robust,tron2009distributed,hartley2013rotation,chatterjee2018robust}, and finally  solve for camera locations \cite{arie2012global,wilson2014robust,ozyesil2015robust,jiang2013global,cui2015linear,cui2015global}. Kasten et al. \cite{kasten2019algebraic} introduced a method for averaging essential matrices, allowing for solving for camera location and orientation in a single optimization framework. In uncalibrated settings, Sweeney et al. \cite{sweeney2015optimizing} presented a method that first improves 
the measured fundamental matrices, and then after self-calibration, apply rotation averaging followed by translation averaging. More recently, \cite{kasten2018gpsfm} introduced an averaging algorithm for fundamental matrices that yields a unique projective reconstruction.

Global methods rely on collections of bifocal tensors, but those cannot determine the magnitudes of translation in collinear camera settings, and 3D points recovered from point tracks in three or more images must be used. This problem has been addressed in the context of translation averaging. Jiang et al. \cite{jiang2013global}   recover translation magnitudes in collinear triplets of cameras by registering the depth of  3D points triangulated independently from each pair of cameras. Akin to our second algorithm, Wilson et al. \cite{wilson2014robust} use   unknown 3D points as additional (but not collinear)  cameras in translation averaging. Cui et al. \cite{cui2015linear}  extend \cite{jiang2013global} to cope with points tracks.

A number of papers analyze the solvability of  SfM by investigating its corresponding viewing graph, in which each node represents a camera and edges represent available fundamental matrices \cite{levi2003viewing,ozyesil2015robust,rudi2010linear,sweeney2015optimizing,trager2018solvability}. These approaches, however, assume that the cameras are in general position and hence do not determine which viewing graphs are solvable in (possibly partly) collinear settings. 

%Recently \cite{kasten2019algebraic} and \cite{kasten2018gpsfm}  characterized the sufficient and necessary conditions for essential and fundamental matrices respectively to be realized by a set of cameras, however their characterization is only valid for a set of cameras which are not all collinear. Finally \cite{sengupta2017new}, derived  a partial list of necessary conditions on the essential and fundamental matrices that are generated from collinear cameras. 

\section{Characterization of collinear settings}

Let $I_1, ..., I_n$ denote a collection of $n$ images of a static scene captured respectively by cameras $P_1,...,P_n$. Each camera $P_i$ is represented by a $3 \times 4$ matrix $P_i=K_i R_i^T[I,-t_i]$ where $K_{i}$ is a $3\times 3$ calibration matrix, $\bf{t}_i\in \mathbb{R}^3$ and $R_i \in SO(3)$ denote the position and orientation of $P_i$, respectively, in some global coordinate system.   We further denote $V_i=K_i^{-T}R_i^{T}$,  therefore the camera projection matrix can be expressed as 
\begin{equation}\label{eq:camera} P_i=V_i^{-T}[I,-{\bf t}_i]\end{equation}  Consequently, let $\bold{X}=(X,Y,Z)^T$ be a scene point in the global coordinate system. Its projection onto $I_i$ is given by ${\bf x}_i = \bold{X}_{i} / Z_i$, where $\bold{X}_{i}=(X_i, Y_i, Z_i)^T = K_{i}R_{i}^{T}(\bold{X}-{\bf t}_{i})$.  %the objective is to retrieve the $n$ positions and rotations of the cameras denoted by ${\bf t}_i \in \Real^3$ and $R_i \in SO(3)$ respectively, where $i=1\dots n$, and are defined up to a global similarity transform of the coordinates system.

We denote the fundamental matrix and the essential matrix between images $I_i$ and $I_j$ by $F_{ij}$ and $E_{ij}$ respectively. It was shown in \cite{arie2012global} that 
$E_{ij}$ and $F_{ij}$ can be written a
\begin{equation}
\label{eq:E}
 E_{ij} = R_i^T (T_i - T_j) R_j \end{equation}
\begin{equation}
\label{eq:F}
 F_{ij} = K_i^{-T} E_{ij} K_j^{-1}=V_i(T_i-T_j)V_j^T\end{equation}
where $T_i=[{\bf t}_i]_{\times}$.

Recently, \cite{kasten2019algebraic,kasten2018gpsfm} established a set of algebraic constraints characterizing the consistency of bifocal tensors for cameras whose center lie in general position. In this paper we complement these characterizations by handling collinear camera centers.   We first repeat the following definitions made in  \cite{kasten2019algebraic,kasten2018gpsfm}. Denote by $\mathbb{S}^{3n}$ 
the set of all the $3n \times 3n$ symmetric matrices.

\begin{definition}\label{def:mv_fundamental} 
A matrix $F \in \mathbb{S}^{3n}$, whose $3 \times 3$ blocks are denoted by $F_{ij}$, is called an {\bf n-view  fundamental matrix} if  $ \forall  i\neq j \in [n],$ $\rank(F_{ij})=2$ and $F_{ii}=0$.  We denote the set of all such matrices by ${\cal F}$.
\end{definition}

\begin{definition}   \label{def:consistent_mv_fundamental} 
An $n$-view  fundamental matrix $F$ is called {\bf consistent}  if there exist camera matrices $P_1,...,P_n$ of the form $P_{i}=V_{i}^{-T}[I,\tr_{i}]$ such that $F_{ij}=V_{i}([\tr_{i}]_{\times}-[\tr_{j}]_{\times})V_{j}^{T}$.
\end{definition}

\begin{definition}  \label{def:mv_essential} 
A matrix $E \in \mathbb{S}^{3n}$, whose $3 \times 3$ blocks are denoted by $E_{ij}$, is called an {\bf n-view essential  matrix} if   $\forall  i\neq j $, $\rank(E_{ij})=2$, the two singular values of $E_{ij}$ are equal, and $E_{ii}=0$. We denote the set of all such matrices by ${\cal E}$.
\end{definition}

\begin{definition}

 An $n$-view essential  matrix $E$ is called {\bf consistent} if there exist $n$ rotation matrices $\{R_i\}_{i=1}^{n}$ and $n$  vectors $\{\tr_i\}_{i=1}^{n}$  such that $E_{ij}= R_{i}^{T}([\tr_{i}]_{\times}-[\tr_{j}]_{\times})R_{j}$. \label{def:consistent_mv_essential}
\end{definition}
We next derive necessary and sufficient conditions for the consistency of essential and fundamental matrices in collinear camera settings.

%\subsection{Calibrated settings}\label{sec:theory_essential}
\begin{theorem}\label{thm:essential} Let $E \in {\cal E}$.
%Let $E \in \mathbb{S}^{3n}$ be an $n$-view essential matrix. 
Then, $E$ is consistent and can be realized by cameras with collinear centers if and only if $E$ satisfies the following two conditions: 
\begin{enumerate}
\item The eigenvalues of $E$ are $\lambda, \lambda, -\lambda, -\lambda$, where $\lambda > 0$.

\item The corresponding eigenvectors, $X, Y \in \Real^{3n \times 2}$, are such that each $3 \times 2$ sub-block, $V_i$, of $\sqrt{0.5}(X+Y)$ satisfies $V_i^T V_i = \frac{1}{n} I_{2 \times 2}$.  
\end{enumerate}
\end{theorem}

%\subsection{Uncalibrated setting}\label{sec:theory_projective}
\begin{theorem}\label{thm:fundamental}
Let $F \in {\cal F}$.
%Let $F \in \mathbb{S}^{3n}$  be an $n$-view fundamental matrix. 
Then, $F$ is   consistent and can be realized by cameras with collinear centers if and only if the following conditions hold
\begin{enumerate}
\item $\rank(F)=4$ and $F$ has exactly 2 positive and 2 negative eigenvalues.
\item $\rank(F_i) = 2,$ where $F_i$ denotes the $i^{th}$ block-row of $F$,  $i \in [n]$.  
\end{enumerate}
\end{theorem}
\noindent The proofs of both theorems are given in the Appendix.
%~\ref{sec:fundamental_proof}.

%-------------------------------------------------------------------------

\section{Method}
In this section we present algorithms for bifocal tensor averaging when either subsets or all of the camera centers are collinear. We assume we are given images $I_1, ..., I_n$ along with a (possibly partial and erroneous) collection of measured bifocal tensors, denoted by $\{ {\hat F}_{ij} \}$ if cameras are uncalibrated or$\{{\hat E}_{ij}\}$ if they are calibrated. Our aim is to find a \textit{consistent} $n$-view bifocal matrix $F \in \mathbb{S}^{3n}$ (resp.~$E \in \mathbb{S}^{3n}$) whose $3 \times 3$ blocks are as close as possible to the measured tensors.

Similar to \cite{kasten2019algebraic,kasten2018gpsfm}, our algorithms rely on constructing a triplet cover of the viewing graph that satisfies certain rigidity-like constraints. Specifically, let $G=(V,W)$ be a viewing graph whose vertices $v_1, ..., v_n \in V$ represent the $n$ cameras, and edges $w_{ij} \in W$ represent pairs of images for which bifocal tensors are measured ($|W| \le {n \choose 2}$).
%In general, only a partial set of the $n \choose 2$ pairwise fundamental (essential) matrices are estimated.  
The information captured in $G$ is summarized in the $n$-view bifocal matrix ${\hat F} \in \mathbb{S}^{3n}$ (resp.\ $\hat E$). 

A triplet cover is a \textit{connected} dual graph $\bar G$, whose nodes represent (possibly a subset of) the 3-cliques in $G$ and edges connect each two vertices whose corresponding  3-cliques in $G$  share an edge (i.e., triplets that share two cameras). Configurations that are represented by such a connected dual graph satisfy a rigidity-like condition, according to which, as is proved in \cite{kasten2018gpsfm} for uncalibrated cameras in general position, if each $9 \times 9$ submatrix of $F$ corresponding to a vertex in $\bar G$ is consistent then $F$ determines the parameters of all cameras uniquely (up to a global projective transformation in ${\cal P}^4$). Moreover, enforcing the consistency of triplets is easier than that of larger sets of cameras since for camera triplets consistency is independent of the scale of the estimated bifocal tensors. Below we denote the number of vertices in $\bar G$ by $m \leq {n \choose 3}$ and index them by $\tau(1),...,\tau(m)$. We further denote by $E_{\tau(k)}$ (resp. $F_{\tau(k)}$) $9 \times 9$ submatrices of $E$ (resp. $F$) corresponding to a triplet $\tau(k)$.

We next present two averaging algorithms. The first algorithm handles image collections taken with cameras whose centers are \textit{all} near collinear. The second algorithm also allows for partial collinearity. We further show how both algorithms can be applied in both calibrated and uncalibrated settings. In each case we formulate the problem as a rank constraint optimization, which we solve using ADMM in a manner similar to \cite{kasten2019algebraic,kasten2018gpsfm}.

%In practice, this is achieved by introducing a novel method that treats points tracks as additional cameras in the viewing graph, allowing a recovery  of  more cameras than in \cite{ kasten2019algebraic} and better accuracy than \cite{kasten2018gpsfm}.   (We note that this approach is also applicable when the entire set of cameras is collinear.)  

\subsection{Fully collinear setups}  \label{sec:rank4}
Our first algorithm applies Thms.~\ref{thm:essential} and~\ref{thm:fundamental} to handle fully collinear setups.   

\subsubsection{Calibrated setting}
Given the measurement matrix ${\hat E}$ and triplet cover $\bar G$ we seek to solve \begin{align} \label{eq:rank4calibrated}
&\underset{E \in {\cal E}}{\min}
& & \sum_{ k= 1}^m ||E_{\tau(k)}-\hat E_{\tau(k)}||^2_F   \\
%& \text{s.t.}& & E=E^T & \nonumber \\ 
%& & &  E_{ii}=0_{3 \times 3}  \nonumber \\
& \text{s.t.} & &  \rank(E_{\tau(k)})=4   \nonumber\\  
& & & \lambda_1(E_{\tau(k)})=-\lambda_4(E_{\tau(k)}),
~\lambda_2(E_{\tau(k)})=-\lambda_3(E_{\tau(k)}) \nonumber 
%\\ & & & \sigma_1(E_{ij})=\sigma_2(E_{ij}),~\sigma_3(E_{ij})=0    \nonumber
\end{align}
where $\lambda_i(.)$ denote the non-zero eigenvalues of a matrix, $i \in [4]$ and $k \in [m]$.  
%The latter constraint enforces  each $3 \times 3$ block to form an essential matrix. 
We note that in \eqref{eq:rank4calibrated} we excluded condition 2 of Thm.\ \ref{thm:essential} to simplify the optimization. Our experiments converged in all cases to solutions that satisfy all the conditions of  Thm.\ \ref{thm:essential}. 
      
\noindent \textbf{Recovery of camera parameters}. Once we obtain an $n$-view essential matrix whose triplets are consistent, we proceed to determine the corresponding $n$ camera matrices. There are two obstacles in this process. First, the obtained essential matrices do not determine the rotations uniquely, and secondly, in collinear settings essential matrices can only determine the direction of the line connecting camera centers, but not positions along this line. Due to the ambiguity of essential matrices three views give rise to eight possible rotation configurations, of which typically four give rise to cyclic consistent configurations (i.e., that satisfy $R_{12}R_{23}R_{31}=I$). To select the appropriate configuration we first use 2-view correspondences as in \cite{hartley2003multiple} to determine the pairwise rotations and then recover the absolute orientation of the three cameras, $R_1,R_2,R_3,$ using the eigenvalues decomposition   method of \cite{arie2012global}.
%Unlike the non-collinear case where each consistent $3$-view essential matrix is realized with  a unique set of $3$ cameras (up to a similarity transformation) \cite{kasten2019algebraic}, in the collinear case, the relative rotations, the magnitudes   and the signs of the relative translations cannot be determined without using the depth information. We solve these ambiguities using points tracks, by first extracting the  cameras from each triplet independently, and then we align all the cameras together.   

%Generally,  for each essential matrix there are $2$ possible  relative rotations \cite{hartley2003multiple}. Therefore, as a result, for $3$ views, there are $8$ possible configurations.  In the case of  non-collinear consistent $3$-view essential matrix,  it is straightforward to select the right configuration, since only  $1$ out\ of the $8$ configurations satisfies the cycle consistency equation $R_{12}R_{23}R_{31}=I$.  However,   for a consistent collinear $3$-view essential matrix, more than one configuration might satisfy this cycle consistency equation. We overcome this ambiguity in the following way. For every pairwise essential matrix, we extract  the right configuration independently using $2$-view    point correspondences as proposed in \cite{hartley2003multiple}. We observe that the resulted configuration of the 3  relative rotations always satisfies the cycle consistency equation.

Next, we need to recover the absolute camera locations. Since our procedure enforces the conditions of Thm.~\ref{thm:essential} for each triplet and due to the rigidity-like structure of the triplet cover, all the recovered essential matrices agree on the direction of the line through the camera center. We therefore set $\tr_1=0$, $\tr_2=-R_{1}\tr_{12}$, and $\tr_3 = \alpha \tr_2 = -\alpha R_1 \tr_{12}$, where the relative translation $\tr_{12}$ is extracted from $\hat E_{12}$ with magnitude 1 and sign determined using 2-view point correspondences. %$t_3=\alpha R_{1}t_{12}=-\alpha t_2$, where $\alpha$ is the scale of $t_{13}$. 
This yields the following camera matrices $P_1 = [R_{1}^T \ |0]$,
$P_2 = [R_{2}^T \ |R_{2}^TR_{1}\tr_{12}]$, and $P_3 = [R_3^T\ | \alpha R_3^TR_1\tr_{12}].$
%\begin{align*}
% P_1 &= [R_{1}^T \ |0] \\
% P_2 &= [R_{2}^T \ |R_{2}^TR_{1}\tr_{12}]\\
% P_3 &= [R_3^T\ | \alpha R_3^TR_1\tr_{12}].
%\end{align*}
To determine $\alpha$ we must resort to 3-view correspondences. Let $\beta_i\xx_i=P_i X$, $i \in [3]$, denote three projections of a 3D point $X$ where $\beta_i$ denote the projective depths of $\xx_i$. As with the  DLT algorithm \cite{hartley2003multiple}, we use the first two equations to determine $X$ and then $\alpha$ is determined from $P_3 X \times \xx_3=0$. Such an equation can be written for every 3-view correspondence, resulting in an over-constrained linear system of equations in $\alpha$ which we solve in least squares. We emphasize that the choice of rotations and $\alpha$ does not change $E$, and so it maintains its consistency and only resolves the ambiguity in reconstructing the underlying cameras. Finally, we use the method in \cite{kasten2019algebraic}, to traversing $\bar G$ and bring all the $n$ cameras to a common Euclidean coordinate frame.

\subsubsection{Uncalibrated setting}
\label{subsec:uncalibrated}

Given measurement matrix ${\hat F}$ and triplet cover $\bar G$ we solve
\begin{align} \label{eq:rank4_fundamental}
& \underset{F \in {\cal \bar F}}{\min}
~~\sum_{ k= 1}^m ||F_{\tau(k)}-\hat F_{\tau(k)}||^2_F   \\
%& \textrm{s.t.}& & F=F^T & \nonumber \\ 
%& & &  F_{ii}=0_{3 \times 3}  \nonumber \\
& \textrm{s.t.}~~~~~ \rank(F_{\tau(k)})=4.   \nonumber  
\end{align}
Here we denote by $\bar{\cal F}$ the set of $n$-view fundamental matrices where we relax the requirement that $\rank(F_{ij})=2$. 
For simplicity of implementation we do not enforce the full set of constraints of Thm.\ \ref{thm:fundamental}. The solutions obtained in our experiments, however, always satisfied all of these conditions.

\noindent \textbf{Recovery of camera parameters}. Once we obtain an $n$-view fundamental matrix whose triplets are consistent we proceed to determine the corresponding $n$ camera matrices.
Here too, due to collinearity, reconstruction is not unique \cite{hartley2003multiple}. Formally, following \cite{hartley2003multiple,levi2003viewing}, given two fundamental matrices ${F}_{12},{F}_{23}$ there are  four degrees of freedom in determine the three camera matrices that are compatible with ${F}_{12}$ and ${F}_{23}$.  
The camera matrices can be expressed as $P_2 = [I \ |0]$, $P_1 = [[\ep_{21}]_{\times}F_{12} \ |\ep_{21}]$, and $P_{3} = [[\ep_{23}]_{\times}F_{23}^T \ |0]+\ep_{23}\mathbf{a}^{T}],$
%\begin{eqnarray*}
% P_2 &=& [I \ |0] \\
% P_1 &=& [[\ep_{21}]_{\times}F_{12} \ |\ep_{21}] \\ 
% P_{3}&=& [[\ep_{23}]_{\times}F_{23}^T \ |0]+\ep_{23}\mathbf{a}^{T}],
%\end{eqnarray*}
where $\ep_{ij}$ is the null-space vector (epipole) of $F_{ij}$, and $\mathbf{a}\in\mathbb{R}^4 $ can be set arbitrarily. For cameras in general position the remaining fundamental matrix ${F}_{13}$ uniquely determines the entries of $\mathbf{a}$. When, however, the three cameras are collinear $\mathbf{a}$ is not determined by ${F}_{13}$. Similar to Sec.\ \ref{sec:rank4}, we  resolve $\mathbf{a}$ using 3-view correspondences. Using the first two view we recover the 3D point $X$ and then obtain equations of the form 
$P_3 X \times \xx_3=0$   which provide two linear equations in $\mathbf{a}$ for every 3-view correspondence.  In principle, two point correspondences suffice to determine $\mathbf{a},$ but for stability we incorporate all 
inlier 3-view correspondences.

This procedure  is applied independently to each triplet of cameras, resulting in camera matrices defined up to a projective transformation. As with the calibrated case, the choice of $\mathbf{a}$ does not change $F$, and it only resolves the ambiguity in reconstructing the cameras. Finally, by traversing $\bar G,$ as in \cite{kasten2018gpsfm}, all the cameras are brought to a common projective coordinate frame.

\subsection{Handling collinearity with virtual cameras}  \label{sec:vc} 

The algorithm presented in Sec.\ \ref{sec:rank4} handles datasets in which \textit{all} cameras are (nearly) collinear. Many common datasets, however, contain both collinear cameras and cameras in general position. We next present a bifocal tensor averaging algorithm that can be applied to any such collection of cameras. Our algorithm extends the averaging algorithms of \cite{kasten2019algebraic,kasten2018gpsfm} to these (partly) accidental settings. The main limitation of those previous algorithms is their reliance on constructing a triplet cover in which \textit{every} triplet must include images captured by cameras in general position. This limits the applicability of the algorithm in datasets that include collinear camera sets and often results in discarding many of the input images. Below we propose a novel approach that overcomes this limitation.

%leverages the strength of triplet of frames, by extending the viewing graph with 3D points that are treated as cameras. This approach allows to  handle  triplets of collinear frames, and to  apply the optimization algorithms of \cite{kasten2018gpsfm,kasten2019algebraic} on scenes which contain collinear segments of frames, yielding a wider coverage of cameras in the scene.  

Our approach is based on augmenting collinear triplets of cameras by constructing virtual cameras centered around 3D points corresponding to 3-view point matches that are not collinear with the real cameras. Let $P_1, P_2$, and $P_3$ be three cameras in a triplet. Recall (Eqs.\  \eqref{eq:camera}-\eqref{eq:F}) that each camera can be parameterized by  $P_i=[V_i^{-T} |- V_i^{-T}\tr_i]\in \mathbb{R}^{3\times 4}$, $i \in [3]$, where in calibrated settings $V_i=R_i^T\in SO(3)$, and the  associated bifocal tensors are then given by   $F_{ij}=V_{i}[\tr_i-\tr_j]_{\times}V_j^T$, $i, j \in [3]$.

Let $X \in \Real^3$ be a 3D point seen by the three cameras. We aim to construct bifocal tensors relating the virtual camera centered at $X$ with the three real cameras $P_1$, $P_2$ and $P_3$. We further choose an ``orientation'' for the virtual camera that coincides with the orientations of one of the real cameras, say, $V_{X}=V_2$.  
The bifocal tensor $F_{iX}$ for $i \in [3]$ can then be expressed as  
\begin{align*}
F_{iX} %&=V_{i}[t_i-X]_{\times}V_X^T \\
& = V_i[\tr_i-X]_{\times} V_2^T\\ 
& = \frac{1}{\det(V_i^{-1})}[V_i^{-T}(\tr_i-X)]_{\times} V_i^{-T}V_2^T,
\end{align*}
where for the latter equality we use the identity  $B^{-1}[{\bf a}]_{\times}=\frac{1}{\det(B)}[B^T {\bf a}]_{\times}B^{T}$.
Let $\xx_i=[x_i, y_i, 1]^T \in \mathbb{R}^{3}$ be the projection of $X$ onto frame $i$. Then it holds that $s_i{\bf x}_i=P_i[X^T, 1]^T=V_i^{-T}(X-\tr_i)$, where $s_i$ is the projective depth of $X$ with respect to camera $i$. Therefore,
\begin{equation}  \label{eq:vf}
F_{iX}=\frac{-s_{i}}{\det(V_i^{-1})}[{\bf x}_i]_{\times} V_{i2},
\end{equation}
where $V_{i2} = V_i^{-T}V_2^T$.
By construction, the matrix
\begin{equation}  \label{eq:virtual4view}
\begin{bmatrix}
   0        & F_{12}  & F_{13} & F_{1X} \\
   F_{12}^T & 0       & F_{23} & F_{2X} \\
   F_{13}^T & F_{23}^T& 0      & F_{3X} \\
   F_{1X}^T & F_{2X}^T& F_{3X}^T &  0 
\end{bmatrix}
\end{equation}
is  a consistent 4-view bifocal matrix. 

Note that $F_{iX}$ in \eqref{eq:vf} can be estimated from the input images, since $V_{i2}$ can be estimated from $\hat F_{12}$ and the scale $-s_i/\det(V_i^{-1})$ can be discarded. Specifically, in a calibrated setting we estimate $V_{i2}=R_{i2}$ from ${\hat E}_{i2}$. Two rotations are obtained, and we use pairwise correspondences to select the correct one.
In an uncalibrated setting, following the recovery of cameras and the use of 3-view correspondences described in Sec.\ \ref{subsec:uncalibrated}, we obtain that $V_{12}=V_1^{-T}V_2=[\ep_{21}]_{\times}F_{12}$, $V_{22}=I$, and $V_{32}=V_3^{-T}V_2=[\ep_{23}]_{\times}F_{23}^T+\ep_{23}[a_{1},a_2,a_3]$.  Finally, $X$ can be selected to be non-collinear with the centers of the three real cameras. Consequently, the estimated elements of \eqref{eq:virtual4view} can be used to augment the viewing graph $G$ and then used in the averaging algorithms of \cite{kasten2019algebraic,kasten2018gpsfm}, which are applicable and stable in general position scenarios. These algorithms use ADMM to solve constrained optimization problems, which for completeness we summarize below. 

{\bf Averaging essential matrices} \cite{kasten2019algebraic}. Given a measurement matrix ${\hat E}$ and triplet cover $\bar G$, we solve 
\begin{align} \label{eq:oprimization_total_objective}
& \underset{E \in {\cal E}}{\min}
~~\sum_{ k= 1}^m ||E_{\tau(k)}-\hat E_{\tau(k)}||^2_F   \\
%& \textrm{s.t.}& & F=F^T & \nonumber \\ 
%& & &  F_{ii}=0_{3 \times 3}  \nonumber \\
& \textrm{s.t.}~~~~~ \rank(E_{\tau(k)})=6   \nonumber \\ 
& ~~~~~~~~~\lambda_i(E_{\tau(k)})=-\lambda_{7-i}(E_{\tau(k)}), ~~ i=1,2,3 \nonumber \\
& ~~~~~~~~~X(E_{\tau(k)})+Y(E_{\tau(k)})~~  {\text {is block rotation,}} \nonumber
\end{align}
where the columns of $X(E_{\tau(k)}), Y(E_{\tau(k)}) \in \Real^{9 \times 3}$ include the eigenvectors of $E_{\tau(k)}$ corresponding respectively to positive and negative eigenvalues.

{\bf Averaging fundamental matrices} \cite{kasten2018gpsfm}. Given a measurement matrix ${\hat F}$ and triplet cover $\bar G$, we solve    
\begin{align} \label{eq:oprimization_total_objective}
& \underset{F \in {\cal \bar F}}{\min}
~~\sum_{ k= 1}^m ||F_{\tau(k)}-\hat F_{\tau(k)}||^2_F   \\
%& \textrm{s.t.}& & F=F^T & \nonumber \\ 
%& & &  F_{ii}=0_{3 \times 3}  \nonumber \\
& \textrm{s.t.}~~~~~ \rank(F_{\tau(k)})=6.   \nonumber  
\end{align}

%In practice,  given a viewing graph, the graph is updated by adding a node per each triplet of cameras that indicates collinearity and in addition shares a triplet correspondence  which is  the projection of some unknown 3D point $X$, and this 3D point is not collinear with these 3 cameras. This node stands for a virtual camera, which is the point location $X$. Therefore, the topology of the viewing graph is modified, in the sense that in addition to the original estimated fundamental matrices, ${\hat F}_{12}, {\hat F}_{13}, {\hat F}_{23}$, we need to have estimations for $$F_{1X}=[{\bf x}_i]_{\times} V_{12},~F_{2X}=[{\bf x}_j]_{\times},F_{3X}=[{\bf x}_k]_{\times} V_{32}.$$
%It is left  to  describe how $V_{i2},~~  i \in [3] $ is computed in the calibrated and  uncalibrated setting.

\section{Experiments}

\begin{table}[tbh]

\caption{\small KITTI, calibrated: Mean position error in meters before BA.  }\label{table:Kitti_Euc}
\resizebox{\linewidth}{!}{
\begin{tabular}{|c||c|c|c|c|c||c|c|c|c|c||c|c|c|c|c|}\hline
\multirow{2}{1.1em}{\textbf{DS}} & \multicolumn{5}{|c||}{ \textbf{5 Cameras}}& \multicolumn{5}{|c||}{ \textbf{10 Cameras}}& \multicolumn{5}{|c|}{\textbf{20 Cameras}} \\\cline{2-16}
& VC &  R4 &  \cite{kasten2019algebraic} &  \cite{ozyesil2015robust} &\cite{wilson2014robust}& VP & R4 & \cite{kasten2019algebraic} &  \cite{ozyesil2015robust} &\cite{wilson2014robust} & VP & R4 &  \cite{kasten2019algebraic} &  \cite{ozyesil2015robust} &\cite{wilson2014robust} \\\hline
 00 
 & \textbf{0.02}
 & {0.06
}
 & {0.36
}
 & 0.05
 &299.50
 & \textbf{0.04
}
 &{ 0.10
}
 & {1.34
}
 & 0.08
 &0.94
 & {0.13
}
 & {0.35
}
 & {2.69
}
 & \textbf{0.11}
 &1.89
 \\\hline
 01 
 & \textbf{0.73
}
 & 1.41
 & {2.36
}
 & 1.16
 &1.88
 & \textbf{1.09
}
 & 3.22
 & {5.47
}
 & 1.86
 &4.26
 & {3.57
}
 &5.21
 & {11.97
}
 & \textbf{2.57}
 &5.87
\\\hline
 02 
 & \textbf{0.03
}
 & {0.04
}
 & 0.65
 & 0.11
 &0.89
 & \textbf{0.09}
 & {0.12
}
 & {1.89
}
 & 0.13
 &1.68
 & 0.34
 & \textbf{0.20
}
 & {4.20
}
 & 0.29
 &1.97
\\\hline
 03 
 & \textbf{0.07
}
 & {0.23
}
 & 0.28
 & 0.09
 &0.55
 & {0.27
}
 & {0.41
}
 & 0.64
 & \textbf{0.08}
 &0.97
 & {0.45
}
 & 1.52
 & 2.33
 & \textbf{0.12
}
 &1.83
\\\hline
 04 
 & \textbf{0.09
}
 & 0.26
 & 0.90
 & 0.18
 &0.84
 & \textbf{0.07}
 & 0.34
 & 2.65
 & 0.18
 &1.74
 & \textbf{0.15
}
 & 0.60
 & {6.22
}
 & 0.39
 &4.21
\\\hline
 05 
 & \textbf{0.02
}
 & {0.06
}
 & 0.49
 & 0.07
 &1166.56
 & \textbf{0.03}
 & {0.10
}
 & {1.37
}
 & 0.13
 &30.14
 & \textbf{0.09
}
 & 0.24
 & {3.24
}
 & 0.19
 &2.02
\\\hline
 06 
 & \textbf{0.03
}
 & {0.16
}
 & {0.94
}
 & 0.14
 &1.37
 & \textbf{0.10
}
 & 0.28
 & {2.43
}
 & 0.17
&1.52
 & {\textbf{0.21}
}
 & 0.76
 & {5.73
}
 & {0.39
}
 &3.89
\\\hline
 07 
 & \textbf{0.02
}
 & 0.08
 & {0.30
}
 & 0.05
&7.95
 & {0.04
}
 & {0.18
}
 & 0.80
 & 0.06
&442.08
 & \textbf{0.09
}
 & 0.34
 & {1.74
}
 & {0.10
}
 &1.79
\\\hline
 08 
 & \textbf{0.02
}
 & {0.04
}
 & {0.40
}
 & 0.06
&0.62
 & \textbf{0.04}
 & \textbf{0.16
}
 & 0.92
 & 0.07
&49.13
 & \textbf{0.11
}
 & 0.25
 & {2.37
}
 & 0.15
 &2.33
\\\hline
 09 
 & \textbf{0.02}
 & {0.15
}
 & {0.62
}
 & 0.12
&21.07
 & {\textbf{0.06
}}
 & {0.24
}
 & 1.96
 & 0.10
 &0.96
 & 0.18
 & 0.53
 & {3.93
}
 & \textbf{0.14
}
 &2.97
\\\hline
 10 
 & \textbf{0.02}
 & {0.04
}
 & 0.61
 & 0.07
& 0.86
 & \textbf{0.05
}
 & 0.14
 & {1.65
}
 & 0.11
&1.42
 & \textbf{0.15}
 & 0.28
 & {3.01
}
 &  {0.17
} &2.54
\\\hline
\end{tabular}
}
\\
%\end{table}
%
%
%
%\begin{table}[tbh]

\caption{\small KITTI, calibrated: average execution time in seconds. }
\label{table:Kitti_euc_time}
\resizebox{\linewidth}{!}{
\begin{tabular}{|c|c|c|c||c|c|c|c||c|c|c|c|}\hline
\multicolumn{4}{|c||}{ \textbf{5 Cameras}}& \multicolumn{4}{|c||}{ \textbf{10 Cameras}}& \multicolumn{4}{|c|}{\textbf{20 Cameras}} \\\cline{1-12}
 VP &  R4 &  \cite{kasten2019algebraic} &  \cite{ozyesil2015robust} & VP & R4 &  \cite{kasten2019algebraic} &    \cite{ozyesil2015robust} & VP & R4 &  \cite{kasten2019algebraic} &  \cite{ozyesil2015robust} \\\hline
  {0.42}
 & 0.47
 & \textbf{0.22}
 & 0.75
 & 1.21
 & 1.07
 & \textbf{0.58}
 & 2.34
 & 3.06
 & 2.15
 & \textbf{1.47}
 & 6.51
 \\\hline
 \end{tabular}
}
%\end{table}
\\
%
%\begin{table}[tbh]

\caption{\small KITTI, uncalibrated: Mean reprojection error  in pixels after BA, averaged per dataset.}\label{table:Kitti_Proj}
\resizebox{\linewidth}{!}{
\begin{tabular}{|c||c|c|c|c||c|c|c|c||c|c|c|c|}\hline
\multirow{2}{1.em}{\textbf{}} & \multicolumn{4}{|c||}{ \textbf{5 Cameras}}& \multicolumn{4}{|c||}{ \textbf{10 Cameras}}& \multicolumn{4}{|c|}{\textbf{20 Cameras}} \\\cline{2-13}
& VC &  R4 &  \cite{kasten2018gpsfm} &  \cite{magerand2017practical} & VP & R4 &  \cite{kasten2018gpsfm} &  \cite{magerand2017practical} & VP & R4 &  \cite{kasten2018gpsfm} &  \cite{magerand2017practical} \\\hline
 00 
 & \textbf{0.12
}
 & \textbf{0.12}
 & {0.40
}
 & 0.13
 & \textbf{0.15}
 & \textbf{0.15}
 & 4.63 & {0.16
}
 &\textbf{ 0.16
}
 &\textbf{ 0.16}
 & {8.71
}
 & 0.18
 \\\hline
 01 
 & 0.90
 & {0.85
} & {2.43
}
 &\textbf{ 0.22}
 & 6.09
 & 1.49
 & 7.09& \textbf{0.32
}
 & 5.84
 & 6.77
 & {16.45
}
 &\textbf{ 0.41}
\\\hline
 02 
 & \textbf{0.12
}
 & \textbf{0.12}
 & 2.84
 & 0.13
 & \textbf{0.15}
 &\textbf{ 0.15}
 & 7.58& {0.16
}
 & \textbf{0.16}
 & \textbf{0.16}
 & {14.20
}
 & 0.19
\\\hline
 03 
 & \textbf{0.14
}
 & \textbf{0.14}
 & 0.91
 & 0.15
 & \textbf{0.17
}
 & \textbf{0.17}
 & 0.84 & 0.21
 & \textbf{0.19
}
 & 0.21
 & 4.60
 & 0.31
\\\hline
 04 
 & \textbf{0.12
}
 &\textbf{ 0.12}
 & 0.69
 & 0.14
 & \textbf{0.15
}
 & \textbf{0.15}
 & 11.68 & 0.18
 & \textbf{0.18}
 & \textbf{0.18}
 & {39.80
}
 & 0.22
\\\hline
 05 
 & \textbf{0.13
}
 & \textbf{0.13}
 & 1.72
 & 0.15
 &\textbf{ 0.16}
 & \textbf{0.16}
 & 9.89 & {0.19
}
 &\textbf{ 0.18}
 & \textbf{0.18}
 & {20.61
}
 & 0.27
\\\hline
 06 
 & \textbf{0.12}
 & \textbf{0.12}
 & {1.92
}
 & 0.13
 &\textbf{ 0.15}
 & \textbf{0.15}
 & 18.74& {0.17
}
 & \textbf{0.17}
 &\textbf{ 0.17}
 & {42.83
}
 & 0.21
\\\hline
 07 
 & 0.84
 & \textbf{0.16}
 & {1.02
}
 & \textbf{0.16}
 & \textbf{0.18
}
 & 0.23
& 3.80 & 0.22
 & \textbf{0.21}
 & 3.15
 & {11.28
}
 & 0.37
\\\hline
 08 
 & \textbf{0.13
}
 & \textbf{0.13
}
 & 0.90
 & 0.14
 & \textbf{0.16
}
 & 0.18
 & 4.16& 0.18
 & \textbf{0.18}
 & 0.21
 & {9.42
}
 & 0.22
\\\hline
 09 
 & \textbf{0.12
}
 & \textbf{0.12}
 & {1.44
}
 & 0.13
 & \textbf{0.14
}
 & \textbf{0.14}
 & 8.78& 0.16
 &\textbf{ 0.16}
 & \textbf{0.16}
 & {12.62
}
 & 0.19
\\\hline
 10 
 & \textbf{0.12
}
 & \textbf{0.12}
 & 1.91
 & 0.13
 & \textbf{0.15}
 & \textbf{0.15}
 & 4.55& {0.17
}
 &\textbf{ 0.17
}
 & 0.20
 & {10.51
}
 &  0.20
\\\hline
\end{tabular}
}
%\end{table}
%
\\
%
%\begin{table}[tbh]

\caption{\small KITTI, uncalibrated: average execution time in seconds }\label{table:Kitti_proj_time}
\resizebox{\linewidth}{!}{
\begin{tabular}{|c|c|c|c||c|c|c|c||c|c|c|c|}\hline
\multicolumn{4}{|c||}{ \textbf{5 Cameras}}& \multicolumn{4}{|c||}{ \textbf{10 Cameras}}& \multicolumn{4}{|c|}{\textbf{20 Cameras}} \\\cline{1-12}
 VP &  R4 &  \cite{kasten2018gpsfm} &  \cite{magerand2017practical} & VP & R4 &  \cite{kasten2018gpsfm} &  \cite{magerand2017practical} & VP & R4 &  \cite{kasten2018gpsfm} &  \cite{magerand2017practical} \\\hline
  \textbf{0.66}
 & 1.24
 & {0.67}
 & 2.04
 & 1.70
 & 2.56
 & \textbf{1.68}
 & 5.60
 & 4.37
 & 5.59
 & \textbf{3.89}
 & 11.63
 \\\hline
 \end{tabular}
}
\end{table}

\begin{table*}[tbh]
\tiny

\caption{\small Unordered internet photos, calibrated: Mean ($\bar x$) and median ($\tilde x$) camera position error in meters before and after bundle adjustment. }
\resizebox{\linewidth}{!}{%

\begin{tabular}{|c|c|c|c|c|c|c|c|c|c|c|c|c|c|c|c|c|c|c|c|c|c|c|}\hline
 & &\multicolumn{5}{|c|}{ \textbf{Our Method } }&\multicolumn{5}{|c|}{ \textbf{ \cite{kasten2019algebraic} } } &    \multicolumn{5}{|c|}{ \textbf{LUD} \cite{ozyesil2015robust}} &\multicolumn{4}{|c|}{ \textbf{ 1DSFM }\cite{wilson2014robust} }  \\\hline
Data set & $N_c$ &$\bar{x}$ & $\tilde{x}$ &$\bar{x}_{BA}$ & $\tilde{x}_{BA}$ & $N_r$ & $\bar{x}$ & $\tilde{x}$ &$\bar{x}_{BA}$ & $\tilde{x}_{BA}$ & $N_r$ & $\bar{x}$ & $\tilde{x}$ &$\bar{x}_{BA}$ & $\tilde{x}_{BA}$ & $N_r$  & $\tilde{x}$ &$\bar{x}_{BA}$ & $\tilde{x}_{BA}$ & $N_r$ \\\hline
Vienna Cathedral & 836&{14.9} &4.9
&  {11.1
} & 2.0
 & 715 & {9.6} & 4.2 & 5.4 & 1.2  & 674 & 10 & 5.4 & 10 & 4.4 & 750  & 6.6 & 2e4 & {0.5} & 757  
\\\hline
Piazza del Popolo & 328& 6.7&3.1
 & 3.2
 & 0.8
 & 280 & 7.2 &  3.5 & {2.5} & {0.8} & 275 & {5} & {1.5} & 4 & 1.0 & 305  & 3.1 & 200 &2.6  & 303  

\\\hline
NYC Library & 332& 4.2 &2.7
  & 2.3
 & 0.8
 
& 281 & {3.3} &  2.2 & {1.1} & 0.47 & 277 & 6 & {2.0} & 7 & 1.4 & 320  & 2.5 & 20 & {0.4} & 292 

\\\hline
Alamo & 577& 2.6&1.1
 & 1.4
 & 0.3
 & 502 & 2.5  & 1.2  & {0.8} & 0.35 & 482 & {2} & {0.4} & 2 & {0.3} & 547 & 1.1 & 2e7 & {0.3} & 521  

 \\\hline

Yorkminster & 437&11.6& 3.1
 & 9.4
 & 1.0
 & 367 & 5.6 & {2.7}  & {1.9} & 0.8 & 341 & {5} & {2.7} & 4 & 1.3 & 404  & 3.4 &  500&{0.2}  & 395 
 
 \\\hline
Montreal ND & 450&2.0 &1  &1.2
 & {0.7
} & 433 & 1.9 &  1.0 & {0.6} & {0.4} & 416 & {1} & {0.5} & 1 & {0.4} & 435  & 2.5 &  1 & 0.9 & 425 
\\\hline
Tower of London & 572& 11.2 &4.6
 & 6.9
 & 1.4
 & 422 & {11.6} & 5.0 & {4} & 1.0 & 414 & 20 & {4.7} & 10 & 3.3 & 425 & 11 &   40&{0.4}  & 414 
 
\\\hline
Ellis Island & 227& 11.0&5.3
 & {4.4
} & 1.8
 & 214 & 14.1 & 6.1 & 5.3 & 1.7 & 211 & - & - & - & - & -  & 3.7 & 40 & {0.4} & 213 

 \\\hline
Notre Dame & 553& {1.7}&0.7 & 0.5
 & {0.2} & 536 & 1.8 & 0.8 & {0.4} & {0.2} & 529 & {0.8} & {0.3} & {0.7} & {0.2} & 536  & 10 & 7&  2.1 & 500  

\\\hline
\end{tabular}
}
\label{table:translation1dsfm}
\end{table*}

\begin{table}[tbh]

{\caption { Unordered internet photos, calibrated: Execution time in seconds. $T_{R+T}$ denote the time for motion averaging (either tensor averaging or rotation and translation). $T_{BA}$ the time for bundle adjustment and $T_{Tot}$ is the total running time of the method, including the additional time for building the triangle cover. Empty cells represent image collections not tested by the authors.}
\label{table:Times}
\resizebox{\linewidth}{!}{%
\
\begin{tabular}{|l|c|c|c|c|c|c|c|c|c|c|c|c|}\hline
&\multicolumn{3}{|c|}{ \textbf{ Our Method } }&\multicolumn{3}{|c|}{ \cite{kasten2019algebraic}} & \multicolumn{3}{|c|}{\textbf{LUD} \cite{ozyesil2015robust}} &\multicolumn{3}{|c|}{ \textbf{1DSFM}\cite{wilson2014robust}}  \\\hline
Data set & $T_{R+T}$ & $T_{BA}$ & $T_{Tot}$ & $T_{R+T}$ & $T_{BA}$ & $T_{Tot}$ &$T_{R+T}$ & $T_{BA}$ & $T_{Tot}$ & $T_{R+T}$ & $T_{BA}$ & $T_{Tot}$  \\\hline
Vienna Cathedral & 145 & 262 & 930& 68 & 293 & 566 & 787 & 208 & 1467 & 323 & 3611 & 3934  \\\hline
Piazza del Popolo & 54 &39 & 143 & 26 & 27 & 87 & 88 & 31 & 162 & 42 & 213 &  255\\\hline
NYC Library & 55 & 80 & 214& 28 & 58 & 125 & 102 & 47 & 200 & 47 & 382 & 429 \\\hline
Alamo & 96 & 115& 509 & 47 & 155 & 327 & 385 & 133 & 750 & 152 &  646 &  798 \\\hline

Yorkminster & 67 & 100 &  296 & 33 & 116 & 207 & 103 & 148 & 297 & 71 & 955 & 1026  \\\hline
Montreal ND & 80 & 216 & 626& 41 & 170 & 494 & 271 & 167 & 553 & 93 & 1043  &  1136 \\\hline
Tower of London & 89 & 132 & 280& 41 & 120 & 241 & 88 & 86 & 228 & 61 & 750 & 811 \\\hline
Ellis Island & 40 & 44 & 170& 21 & 53 & 140 & - & - & - & 29 & 276 &  305  \\\hline
Notre Dame & 100 & 419 & 1070 & 52 & 277 & 720& 707 & 126 & 1047 & 205 & 2139 &  2344  \\\hline
\end{tabular}
}}
 \label{table:translation}
\end{table}

\begin{table}[tbh]

\caption{\small Unordered internet photos, uncalibrated: Mean reprojection error  and execution times. $m$ and $n$ respectively denote the number of 3D points and cameras.}\label{table:Olsson}
\resizebox{\linewidth}{!}{%
\
\begin{tabular}{|l|r|r|rrr|rrr|}
\hline
 \multirow{2}{2em}{\textbf{Dataset}} & \multirow{2}{1em} {\textbf{m}} &\multirow{2}{1em}{\textbf{n}} &\multicolumn{3}{|c|}{ \textbf{Error (pixels)} }  & \multicolumn{3}{|c|}{\textbf{Time (sec.)} }  \tabularnewline
\cline{4-9}
 &  &  & \textbf{VC} & \textbf{\cite{kasten2018gpsfm}} & \textbf{\textbf{\cite{magerand2017practical}}}   &\textbf{VC} & \textbf{\cite{kasten2018gpsfm}} &\textbf{\cite{magerand2017practical}}  \tabularnewline
\hline
%Dino 319 & 319 & 36 &  \textbf{0.43} &\textbf{0.43} & 0.47  &  {12.32} &{3.64} & \textbf{3.46}  \tabularnewline
Dino 4983 & 4983 & 36 & {0.43} &\textbf{0.42} & 0.47  &{15.75} &\textbf{4.65} & 13.00   \tabularnewline
%Corridor & 737 & 11 & \textbf{0.26} &\textbf{0.26}  & 0.27 & {2.48} &\textbf{1.03} & 1.55  \tabularnewline
%House & 672 & 10 & \textbf{0.34} & \textbf{0.34} & 0.40  & 1.82 & \textbf{0.94}& {1.03} \tabularnewline
%Gustav Vasa & 4249 & 18 & \textbf{0.16} & \textbf{0.16} & 0.17 &   {4.90} & \textbf{2.47} & 6.64 \tabularnewline
Folke Filbyter & 21150 & 40 & \textbf{0.26} & {0.82} & 0.31 & {14.30} & \textbf{6.70} &  102.77 \tabularnewline
%Park Gate & 9099 & 34 & \textbf{0.31} &\textbf{0.31} & 0.45 & {19.68}& \textbf{9.25} & 31.58 \tabularnewline
%Nijo & 7348 & 19 & \textbf{0.39} &\textbf{0.39} & 0.44 & {7.02}&\textbf{3.80}& 12.68 \tabularnewline
%Drinking Fountain & 5302 & 14 & \textbf{0.28}& \textbf{0.28} & 0.31 & {4.64} & \textbf{2.12} &9.37 \tabularnewline
%Golden Statue & 39989 & 18 & \textbf{0.22}& \textbf{0.22} & 0.23 & {10.08} & \textbf{5.05} & 36.21 \tabularnewline
%Jonas Ahls & 2021 & 40 & \textbf{0.18}& \textbf{0.18} & 0.20  & {13.84}& \textbf{5.49} & 13.40 \tabularnewline
%De Guerre & 13477 & 35 & \textbf{0.26} &\textbf{0.26} & 0.28 & {34.32} &\textbf{11.19}& 32.67 \tabularnewline
%Dome & 84792 & 85 & \textbf{0.24} & \textbf{0.24} & 0.25 &{108.18}& \textbf{65.12} & 226.13 \tabularnewline
%Alcatraz Courtyard & 23674 & 133 & \textbf{0.52} & \textbf{0.52} & 0.57 & {126.40}& \textbf{63.94} & 151.28 \tabularnewline
%Alcatraz Water Tower & 14828 & 172 & \textbf{0.47} & \textbf{0.47} & 0.59 &  169.08 &  90.24 & \textbf{71.80} \tabularnewline
Cherub & 72784 & 65 & {0.75} & \textbf{0.74}& 0.81   &{48.52} &\textbf{27.30}& 101.64 \tabularnewline
%Pumpkin & 69335 & 195 & \textbf{0.38} & \textbf{0.38}  &0.44  & {203.06} &\textbf{93.32} & 222.09\tabularnewline
%Sphinx & 32668 & 70 & \textbf{0.34}& \textbf{0.34}& 0.36 &  {39.63} & \textbf{31.41} &79.91 \tabularnewline
Toronto University & 7087 & 77 & \textbf{0.24} & 0.54& 0.26  & {30.47}& \textbf{26.59} & 91.26 \tabularnewline

Sri Thendayuthapani & 88849 & 98 & \textbf{0.31} & 0.51 & 0.33   & \textbf{219.11} &{220.25} & 325.58 \tabularnewline
 %Porta san Donato & 25490 & 141 & \textbf{0.40} & \textbf{0.40} &3.56 & {126.54} & \textbf{82.18} & 157.96 \tabularnewline
%Buddah Tooth & 27920 & 162 & \textbf{0.60} & \textbf{0.60} & 0.71 &  {142.02} & \textbf{59.75} & 81.05 \tabularnewline
Tsar Nikolai I & 37857 & 98 & \textbf{0.29}  & {0.32} & 0.31 
& {89.93} & \textbf{70.79} & 101.01 \tabularnewline
Smolny Cathedral & 51115 & 131 & \textbf{0.46} & {0.48} &  0.50  & {303.62} & \textbf{210.75} & 263.60\tabularnewline
Skansen Kronan & 28371 & 131 & \textbf{0.41} & 0.44 & 0.44  &{118.60} &\textbf{83.43} &161.81 \tabularnewline
\hline
\end{tabular}
}
\label{table::sfm_uncalibrated}
\end{table}

We evaluate our algorithms on several datasets, including nearly collinear video sequences taken from the KITTI Visual Odometry datasets \cite{Geiger2012CVPR} and unordered collections of calibrated \cite{wilson2014robust} and uncalibrated \cite{olsson2011stable,vgg_dataset} images. Since our first algorithm (Sec.~\ref{sec:rank4}), which we denote by `R4,' is applicable only to fully collinear settings we apply it only to the KITTI sequences. The second algorithm (Sec.~\ref{sec:vc}), denoted `VC,' is applied to all datasets. We compare our algorithms to several recent methods, including\cite{kasten2019algebraic} , LUD~\cite{ozyesil2015robust}  and 1DSFM~\cite{wilson2014robust} for the  calibrated datasets and GPSFM \cite{kasten2018gpsfm} and PPSFM~ \cite{magerand2017practical}  for the uncalibrated ones. For the calibrated settings we compare the mean and median translation errors and for non-calibrated settings we compare the mean reprojection error. 

\subsection{Datasets}

\noindent\textbf{Driving Car Image Collections}.
The KITTI visual odometry \cite{Geiger2012CVPR} benchmark includes 11 video sequences captured by moving cars with ground truth camera positions and orientations. As is typical for driving, these sequences often contain stretches of near collinear motion. For our experiments we randomly selected for each of 11 datasets three near collinear subsequences (identified by applying PCA to the ground truth camera locations), each includes 100 frames. We then used each sequence to produce three collections of non-overlapping subsequences, each of length 5, 10 or 20 frames, yielding a total of 1155 sequences over 3300 frames.

\noindent\textbf{Unordered Internet Photos}.
We further test our VC algorithm on calibrated unordered internet photo collections, collected by \cite{wilson2014robust}. We note that the ``ground truth" camera matrices for this dataset include an estimate obtained by an incremental method \cite{snavely2006photo} .
This dataset includes many outlier photographs, and consequently, in addition to maximizing accuracy, our goal is to maximize the number of cameras handled by the method. Additional datasets include uncalibrated photos \cite{olsson2011stable,vgg_dataset}. As ``ground truth" these datasets include a list of inliers 2D  projections  of  unknown  3D points by unknown cameras, allowing us to evaluate reconstruction accuracy via the mean reprojection error.

\subsection{Constructing a triplet cover}

Both our R4 and VC algorithms require a triplet cover graph $\bar G$ as input. We produce this cover by applying the following three steps. 

\noindent{\textbf{Initial triplet cover}}.
For our R4 algorithm, we initialize $\bar G$ simply using consecutive camera triplets $(i-1,i,i+1)$, $2 \le i <n-1$. 
For our VC algorithm we construct an initial cover using the heuristics of \cite{kasten2019algebraic,kasten2018gpsfm}, where for the Kitti datasets we do not  filter collinear triplets.

\noindent{\textbf{Enrichment}}. For the internet photo collection datasets we next enrich the initial triplet covers as follows. For the calibrated dataset of \cite{wilson2014robust}, the initial cover is typically disconnected. Rather than using just the largest connected component, as in \cite{kasten2019algebraic}, we augment the set of triplets with collinear ones that make the graph connected. To that end, we also keep the full cover $\bar G'$, a graph that includes all the 3-cliques in $G$ (so the vertices of $\bar G'$ form a superset of the vertices of $\bar G$). We then iteratively select pairs of nodes in $\bar G$ from different connected components and use shortest path (measured by the number of edges) to connect them in $\bar G'$. 

For the uncalibrated datasets \cite{olsson2011stable,vgg_dataset} the initial triplet cover produced with \cite{kasten2018gpsfm} forms graphs with single connected components. Although the triplets in these graphs pass the non-collinearlity test applied in \cite{kasten2018gpsfm}, we add a virtual camera to each triplet, in order to improve the stability of the averaging algorithm.

\noindent{\textbf{Adding virtual cameras and filtering}}. Next, we identify collinear triplets (using the collinearity  measures of \cite{kasten2019algebraic,kasten2018gpsfm})   and for each such triplet produce a virtual camera. To that end, we consider the set of 3-view correspondences that are not collinear with the 3 camera centers (we avoid points whose too close to the epipole), and select the match that minimizes the sum of symmetric epipolar distances in all three images. We then produce the three bifocal tensors relating the virtual camera to the triplet. As this results in a cover that is not minimal (for example, only two of the three triplets that involve a virtual camera are needed to produce a valid cover), we further remove superfluous triplets from cover as in \cite{kasten2019algebraic,kasten2018gpsfm}.

%In the projective case, there is no need to connect connected component because the triplet graph of \cite{kasten2018gpsfm} already cover all the cameras.  However, for some of the cases, in order to keep their graph connected their threshold is relatively permissive. As a result their  optimization still suffers from inaccuracies that are resulted from  sensitivity to the remaining collinear triplets. To this end, we take the final graph of \cite{kasten2018gpsfm} and replace by two triplets  resulted from adding a point, all the triplets. The replacement is done in the same as in \ref{}

\subsection{Results}

Tables \ref{table:Kitti_Euc}-\ref{table::sfm_uncalibrated} show the accuracy and execution times obtained with our algorithms on all datasets.
Our algorithms are further compared to state of the art methods including essential matrix averaging \cite{kasten2019algebraic}, LUD \cite{ozyesil2015robust}, and 1DSFM \cite{wilson2014robust} for calibrated images and GPSFM \cite{kasten2018gpsfm} and PPSFM \cite{magerand2017practical} for uncalibrated ones.
Tables \ref{table:Kitti_Euc}-\ref{table:Kitti_proj_time} 
%\ref{table:Kitti_euc_time}, and \ref{table:Kitti_Proj} 
show the results of applying our algorithms on the nearly collinear Kitti sequences under calibrated and uncalibrated settings. These experiments demonstrate that our pipelines are faster than the other methods except \cite{kasten2019algebraic,kasten2018gpsfm}, which completely fail to reconstruct the scene due to their sensitivity to collinearity. Also, our algorithms were more accurate than the other methods in most of the runs.
%In the calibrated case, out of eleven datasets, our methods achieved the best results in 11,9,7 cases from the 5,10,20 cameras respectively. For the uncalibrated settings, our of eleven datasets we achieved the best results in ten at each of the 5,10,20 cameras cases.

Tables \ref{table:translation1dsfm}-\ref{table:Times} show the results of applying our VC algorithm to the calibrated internet photo collection \cite{wilson2014robust}. The results demonstrate the benefit of adding virtual cameras for collinear triplets, which has led to increasing the number of reconstructed cameras compared to \cite{kasten2019algebraic} while maintaining comparable accuracy. Our method runs as fast as LUD and is much faster than 1DSFM.

Table \ref{table::sfm_uncalibrated} shows reprojection error and execution time of our VC algorithm on the uncalibrated internet photo\ benchmark. Compared to GPSFM and PPSFM, our reconstruction is more accurate than the other methods in 
6 out of the 8 datasets and is on par with GPSFM on the remaining 17 datasets.
Our method is slower than GPSFM due to the additional virtual cameras.
Yet it is  faster than PPSFM in most runs and more accurate in all runs. 
Additional results are provided in the supplementary material. 

\noindent\textbf{Technical details}.
We ran our experiments on an Intel(R)-i7 3.20GHz with Windows. For BA we used the   Theia SfM library~\cite{theia-manual}, which we run on a Linux Intel(R) Xeon(R) CPU@2.30GHz with 16 cores. Camera position results for \cite{kasten2019algebraic,ozyesil2015robust,wilson2014robust} in Tables \ref{table:translation1dsfm}-\ref{table:Times} are taken from these papers.

\begin{comment}
\subsubsection{Graph construction - \yk{not ready }}
Next for a viewing graph that contains the 3 fundamental matrices $F_{12},F_{13},F_{23}$ of a near  collinear cameras:    the  connections $F_{1x},F_{2x},F_{3x}$ are  added . We note that since cameras $1,2,3 $ are near collinear, they could not be optimizes to rank 6 by the optimization algorithm of \cite{gpsfm}, and such triplet were filtered out of the triplet cover in their algorithm. Here, instead of filtering such a triplet we suggest to replace it by two triplets that ivolves two of the 3 original fundamental matrices and the 3 new fundamental matrices. This way, the triplet cover of the viewing graph,  can use the near collinear triplet and optimize directly for cameras $1,2,3. $ and fveer $X$. 
We  make a slight change to the triplet cover construction that we will describe next.\\
For the graph construction, we consider a weighted viewing graph $G$ where the weight for each edge $w_{ij}$ is assigned to be the number of inliers between the corresponding frames. We begin be removing a disjoint set of minimal spanning trees from them we generate the intial triplets collection. Similarly to \cite{cite some one who does it} we remove inconsistent triplet by their rotation consistency score. 
\end{comment}

{\small \noindent\textbf{Acknowledgment} This research is supported in part by the U.S.-Israel Binational Science Foundation, grant number 2018680, by the Minerva Foundation with funding from the German Federal Ministry for Education and Research, and by the Kahn foundation. }

\clearpage
{\small
\bibliographystyle{ieee_fullname}
\bibliography{egbib}

\begin{thebibliography}{10}\itemsep=-1pt

\bibitem{agarwal2009building}
Sameer Agarwal, Noah Snavely, Ian Simon, Steven~M Seitz, and Richard Szeliski.
\newblock Building rome in a day.
\newblock In {\em 2009 IEEE 12th international conference on computer vision},
  pages 72--79. IEEE, 2009.

\bibitem{arie2012global}
Mica Arie-Nachimson, Shahar~Z Kovalsky, Ira Kemelmacher-Shlizerman, Amit
  Singer, and Ronen Basri.
\newblock Global motion estimation from point matches.
\newblock In {\em 2012 Second Int. Conf. on 3D Imaging, Modeling, Processing,
  Visualization \& Transmission}, pages 81--88. IEEE, 2012.

\bibitem{camposeco2018hybrid}
Federico Camposeco, Andrea Cohen, Marc Pollefeys, and Torsten Sattler.
\newblock Hybrid camera pose estimation.
\newblock In {\em Proceedings of the IEEE Conference on Computer Vision and
  Pattern Recognition}, pages 136--144, 2018.

\bibitem{chatterjee2018robust}
Avishek Chatterjee and Venu~Madhav Govindu.
\newblock Robust relative rotation averaging.
\newblock {\em IEEE Transactions on Pattern Analysis and Machine Intelligence},
  40(4):958--972, 2018.

\bibitem{cui2015linear}
Zhaopeng Cui, Nianjuan Jiang, Chengzhou Tang, and Ping Tan.
\newblock Linear global translation estimation with feature tracks.
\newblock {\em ArXiv preprint, ArXiv:1503.01832}, 2015.

\bibitem{cui2015global}
Zhaopeng Cui and Ping Tan.
\newblock Global structure-from-motion by similarity averaging.
\newblock In {\em Proceedings of the IEEE International Conference on Computer
  Vision}, pages 864--872, 2015.

\bibitem{gao2003complete}
Xiao-Shan Gao, Xiao-Rong Hou, Jianliang Tang, and Hang-Fei Cheng.
\newblock Complete solution classification for the perspective-three-point
  problem.
\newblock {\em IEEE transactions on pattern analysis and machine intelligence},
  25(8):930--943, 2003.

\bibitem{Geiger2012CVPR}
Andreas Geiger, Philip Lenz, and Raquel Urtasun.
\newblock Are we ready for autonomous driving? the kitti vision benchmark
  suite.
\newblock In {\em Conference on Computer Vision and Pattern Recognition
  (CVPR)}, 2012.

\bibitem{hartley2013rotation}
Richard Hartley, Jochen Trumpf, Yuchao Dai, and Hongdong Li.
\newblock Rotation averaging.
\newblock {\em International Journal of Computer Vision}, 103(3):267--305,
  2013.

\bibitem{hartley2003multiple}
Richard Hartley and Andrew Zisserman.
\newblock {\em Multiple view geometry in computer vision}.
\newblock Cambridge University Press, 2003.

\bibitem{jiang2013global}
Nianjuan Jiang, Zhaopeng Cui, and Ping Tan.
\newblock A global linear method for camera pose registration.
\newblock In {\em IEEE Int. Conf. on Computer Vision}, pages 481--488, 2013.

\bibitem{kasten2019resultant}
Yoni Kasten, Meirav Galun, and Ronen Basri.
\newblock Resultant based incremental recovery of camera pose from pairwise
  matches.
\newblock In {\em 2019 IEEE Winter Conference on Applications of Computer
  Vision (WACV)}, pages 1080--1088. IEEE, 2019.

\bibitem{kasten2019algebraic}
Yoni Kasten, Amnon Geifman, Meirav Galun, and Ronen Basri.
\newblock Algebraic characterization of essential matrices and their averaging
  in multiview settings.
\newblock In {\em The IEEE International Conference on Computer Vision (ICCV)},
  October 2019.

\bibitem{kasten2018gpsfm}
Yoni Kasten, Amnon Geifman, Meirav Galun, and Ronen Basri.
\newblock Gpsfm: Global projective sfm using algebraic constraints on
  multi-view fundamental matrices.
\newblock In {\em The IEEE Conference on Computer Vision and Pattern
  Recognition (CVPR)}, June 2019.

\bibitem{klopschitz2010robust}
Manfred Klopschitz, Arnold Irschara, Gerhard Reitmayr, and Dieter Schmalstieg.
\newblock Robust incremental structure from motion.
\newblock In {\em Proc. 3DPVT}, volume~2, pages 1--8, 2010.

\bibitem{levi2003viewing}
Noam Levi and Michael Werman.
\newblock The viewing graph.
\newblock In {\em Computer Vision and Pattern Recognition}, volume~1, pages
  I--I. IEEE, 2003.

\bibitem{magerand2017practical}
Ludovic Magerand and Alessio Del~Bue.
\newblock Practical projective structure from motion (p2sfm).
\newblock In {\em 2017 IEEE Int. Conf. on Computer Vision (ICCV)}, pages
  39--47. IEEE, 2017.

\bibitem{martinec2007robust}
Daniel Martinec and Tomas Pajdla.
\newblock Robust rotation and translation estimation in multiview
  reconstruction.
\newblock In {\em 2007 IEEE Conference on Computer Vision and Pattern
  Recognition}, pages 1--8. IEEE, 2007.

\bibitem{olsson2011stable}
Carl Olsson and Olof Enqvist.
\newblock Stable structure from motion for unordered image collections.
\newblock In {\em Scandinavian Conf. on Image Analysis}, pages 524--535.
  Springer, 2011.

\bibitem{ozyesil2015robust}
Onur Ozyesil and Amit Singer.
\newblock Robust camera location estimation by convex programming.
\newblock In {\em IEEE Conf. on Computer Vision and Pattern Recognition}, pages
  2674--2683, 2015.

\bibitem{persson2018lambda}
Mikael Persson and Klas Nordberg.
\newblock Lambda twist: an accurate fast robust perspective three point (p3p)
  solver.
\newblock In {\em Proceedings of the European Conference on Computer Vision
  (ECCV)}, pages 318--332, 2018.

\bibitem{pollefeys2004visual}
Marc Pollefeys, Luc Van~Gool, Maarten Vergauwen, Frank Verbiest, Kurt Cornelis,
  Jan Tops, and Reinhard Koch.
\newblock Visual modeling with a hand-held camera.
\newblock {\em Int. Journal of Computer Vision}, 59(3):207--232, 2004.

\bibitem{rudi2010linear}
Alessandro Rudi, Matia Pizzoli, and Fiora Pirri.
\newblock Linear solvability in the viewing graph.
\newblock In {\em Asian Conf. on Computer Vision}, pages 369--381. Springer,
  2010.

\bibitem{schonberger2016structure}
Johannes~L Schonberger and Jan-Michael Frahm.
\newblock Structure-from-motion revisited.
\newblock In {\em Proceedings of the IEEE Conference on Computer Vision and
  Pattern Recognition}, pages 4104--4113, 2016.

\bibitem{sengupta2017new}
Soumyadip Sengupta, Tal Amir, Meirav Galun, Tom Goldstein, David~W Jacobs, Amit
  Singer, and Ronen Basri.
\newblock A new rank constraint on multi-view fundamental matrices, and its
  application to camera location recovery.
\newblock In {\em Proceedings of the IEEE Conference on Computer Vision and
  Pattern Recognition}, pages 4798--4806, 2017.

\bibitem{snavely2006photo}
Noah Snavely, Steven~M Seitz, and Richard Szeliski.
\newblock Photo tourism: exploring photo collections in 3d.
\newblock In {\em ACM transactions on graphics (TOG)}, volume~25, pages
  835--846. ACM, 2006.

\bibitem{snavely2008modeling}
Noah Snavely, Steven~M Seitz, and Richard Szeliski.
\newblock Modeling the world from internet photo collections.
\newblock {\em International Journal of Computer Vision}, 80(2):189--210, 2008.

\bibitem{theia-manual}
Chris Sweeney.
\newblock Theia multiview geometry library: Tutorial \& reference.
\newblock \url{http://theia-sfm.org}.

\bibitem{sweeney2015optimizing}
Chris Sweeney, Torsten Sattler, Tobias Hollerer, Matthew Turk, and Marc
  Pollefeys.
\newblock Optimizing the viewing graph for structure-from-motion.
\newblock In {\em IEEE Int. Conf. on Computer Vision}, pages 801--809, 2015.

\bibitem{trager2018solvability}
Matthew Trager, Brian Osserman, and Jean Ponce.
\newblock On the solvability of viewing graphs.
\newblock In {\em European Conf. on Computer Vision}, pages 335--350. Springer,
  Cham, 2018.

\bibitem{triggs1999bundle}
Bill Triggs, Philip~F McLauchlan, Richard~I Hartley, and Andrew~W Fitzgibbon.
\newblock Bundle adjustment—a modern synthesis.
\newblock In {\em Int. Workshop on Vision Algorithms}, pages 298--372.
  Springer, 1999.

\bibitem{tron2009distributed}
Roberto Tron and Ren{\'e} Vidal.
\newblock Distributed image-based 3-d localization of camera sensor networks.
\newblock In {\em Proceedings of the 48h IEEE Conference on Decision and
  Control (CDC) held jointly with 2009 28th Chinese Control Conference}, pages
  901--908. IEEE, 2009.

\bibitem{vgg_dataset}
Oxford VGG.
\newblock Multiview datasets.
\newblock \url{http://www.robots.ox.ac.uk/~vgg/data/}.

\bibitem{wilson2014robust}
Kyle Wilson and Noah Snavely.
\newblock Robust global translations with 1dsfm.
\newblock In {\em European Conf. on Computer Vision}, pages 61--75. Springer,
  2014.

\bibitem{wu2013towards}
Changchang Wu.
\newblock Towards linear-time incremental structure from motion.
\newblock In {\em 2013 International Conference on 3D Vision-3DV 2013}, pages
  127--134. IEEE, 2013.

\bibitem{zheng2015structure}
Enliang Zheng and Changchang Wu.
\newblock Structure from motion using structure-less resection.
\newblock In {\em Proceedings of the IEEE International Conference on Computer
  Vision}, pages 2075--2083, 2015.

\end{thebibliography}
}
\begin{appendices}

\onecolumn
\section*{Appendix}

\renewcommand{\thesubsection}{\Alph{subsection}}

%Our proofs
%rely on lemmas provided in the supplementary material.

\section{Proof of Theorem~\ref{thm:essential}} \label{sec:essential_proof}

\begin{proof}$\Rightarrow:$ Assume $E \in {\cal E}$ is consistent and realized by cameras with collinear centers.  Using Thm. 2 %\ref{thm:fundamental}
in  \cite{kasten2019algebraic}
we have 
\begin{enumerate}[i.]
\item $E$ can be formulated as $E = A + A^T$ where $A=U V^T$ and $U, V \in \Real^{3n \times 3}$
\begin{align}  \label{eq:uv}
U =\left[\begin{array}{c}
\alpha_1 R_{1}^T \\
\vdots\\
\alpha_n R_{n}^T
\end{array}\right]T, ~~ & ~~
V =\left[\begin{array}{ccc}
R_{1}^T\\
\vdots\\
R_{n}^T
\end{array}\right]
\end{align}
with $R_i \in SO(3)$, ${\bf t}_i = \alpha_i \tr$, $i \in [n]$,  $\tr \in \Real^3$, $T = [\tr]_{\times}$, such that $ \sum_{i=1}^n \tr_i=\sum_{i=1}^n \alpha_i \tr = 0$.

\item Each column of $U$ is orthogonal to each column of $V$, i.e., $V^T U = 0_{3 \times 3}$.
%\item $\rank(V)=3$
\end{enumerate}
To prove condition 1, we first examine the eigenvalues of
\begin{equation}  \label{eq:ata}
A^TA=VU^TUV^{T}.
\end{equation}
Due to the orthogonality of the $R_i$'s
\begin{equation}
U^TU = \left(\sum_{i=1}^n \alpha_i^2\right) T^TT.
\end{equation}
% \begin{align*}
% U^TU &= -T
% \begin{bmatrix}
% \alpha_1 R_1, \ldots, \alpha_n R_n
% \end{bmatrix}
% \begin{bmatrix}
% \alpha_1 R_1^T \\ \vdots \\ \alpha_n R_n^T
% \end{bmatrix}
% T \\
% &=-T (\sum_{i=1}^n \alpha_i^2) I_{3 \times 3} T \\
% &=\sum_{i=1}^n \alpha_i^2 (T^TT).
% \end{align}
$T$ is $3 \times 3$ skew-symmetric, therefore it has 2 identical singular values (while the third one is zero). Consequently, the (symmetric) matrix $U^TU$ has two identical eigenvalues, with eigenvalue decomposition of the form $U^T U = P \tilde\Lambda P^T$ with $\tilde\Lambda=\diag[\tilde\lambda,\tilde\lambda,0]$ and $P \in SO(3)$. Therefore $\rank(U) = 2$ and, using \eqref{eq:ata},
\begin{equation}
\label{eq:eigen_decomp}
A^T A = VP \tilde\Lambda P^TV^T.
\end{equation}

Note that $(VP)^T (VP) = P^TV^TVP = n I_{3 \times 3}$, therefore  if we scale $V$ by  $1/\sqrt{n}$ and scale $\tilde\lambda$ by $n$ \eqref{eq:eigen_decomp} becomes the eigen-decomposition of $A^TA$. Thus, $A^T A$ has two equal eigenvalues, and since $\lambda(A^TA) = (\sigma(A))^2$ it follows that  $A$ has two equal singular values (and $\rank(A) = 2$).  Consequently, the full \rank\
factorization of $A$ is of the form $A = {\tilde U} {\tilde V}^T$, where $\tilde{U}, \tilde{V} \in \Real^{3n \times 2}$.

 Next, we  construct a full-rank factorization of $A$ based on the columns of  $U, V \in \Real^{3n \times 3}$. Based on the observation that $\rank(U)=2$,  without loss of generality we assume that the third column of $U$ is a linear combination of the first two columns, i.e., $\uu_3 = \alpha \uu_1 + \beta \uu_2$ for some $\alpha, \beta \in \Real$. Now, $A$ can be decomposed as
$A = U V^T = \sum_{i=1}^3 \uu_i \vv_i^T = \uu_1(\vv_1^T+\alpha \vv_3^T)+\uu_2(\vv_2^T +\beta \vv_3^T).$
%\begin{align*}
%A & = U V^T\\
%&= \sum_{i=1}^3 \uu_i \vv_i^T \\
%&= \sum_{i=1}^2 \uu_i \vv_i^T + (\alpha \uu_1 + \beta \uu_2) v_3^T \\
%& = \uu_1(\vv_1^T+\alpha \vv_3^T)+\uu_2(\vv_2^T +\beta \vv_3^T).
%\end{align*}
So a full-rank decomposition of $A$ takes the form $A = {\tilde U} {\tilde V}^T$, with $\tilde U = [\uu_1, \uu_2]$ and $\tilde V = [\vv_1+\alpha \vv_3, \vv_2+\beta \vv_3]$. Note that ${\text {span}}(\tilde U) = {\text {span}}(U)$ and ${\text {span}}(\tilde V) \subseteq {\text {span}}(V)$, implying that the columns of $\tilde U$ are orthogonal to the columns of $ \tilde V$.

Let $A = {\hat U} \begin{bmatrix} \sigma & 0 \\ 0 & \sigma \end{bmatrix} {\hat V}^T$ be the  SVD decomposition of $A$, where ${\hat U}, {\hat V} \in \Real^{3n \times 2}$, then
\begin{equation}
\label{eq:E_svd}
E = A + A^T = \begin{bmatrix} {\hat U} & {\hat V}  \end{bmatrix} \Sigma \begin{bmatrix} {\hat V}^T \\ {\hat U}^T\end{bmatrix},
\end{equation}
where $\Sigma=\sigma I \in \mathbb{S}^4$. We next claim that  \eqref{eq:E_svd} is the SVD decomposition of $E$. To that end, we need to show that ${\text {span}}({\hat U})$ and ${\text {span}}({\hat V})$ are orthogonal sub-spaces.   Recall that $A$ can be factorized into $A={\tilde U} {\tilde V}^T$, such that ${\text {span}}(\tilde U)$ and ${\text {span}}(\tilde V)$ are orthogonal sub-spaces.   Due to the ambiguity of the full-rank  factorization, there exists an invertible matrix $B \in \Real^{2 \times 2}$ such that ${\hat U} = \tilde U B$ and ${\hat V} = \tilde V B^{-T}$  which implies that   ${\text {span}}({\hat U})$  is orthogonal to ${\text {span}}({\hat V})$. This derivation implies that $\rank(E) =4$ with 4 identical singular values.  Finally, using Lemma ,\ref{lemma:SVDtoSPEC} 
due to the form of singular vectors the eigenvalues of $E$ are of the form $\lambda, \lambda, -\lambda$, $-\lambda$, where $\lambda=\sigma > 0$.

Next, to prove the second condition, we first construct an SVD decomposition of $A$, using $A = UV^T$ in (i) and the  thin SVD decomposition  of  $T = P \begin{bmatrix} \sigma & 0 \\ 0 & \sigma \end{bmatrix} Q^T$ with  $P, Q \in \Real^{3 \times 2}$ such that $P^T P = Q^T Q = I_{2 \times 2}$.
Let  ${\hat U} = \frac{1}{\sqrt{\sum {\alpha_i^2}}} \begin{bmatrix} \alpha_1 R_1^T \\ \vdots \\ \alpha_n R_n^T\end{bmatrix} P$ and ${\hat V} = \frac{1}{\sqrt{n}} \begin{bmatrix}  R_1^T \\ \vdots \\  R_n^T\end{bmatrix} Q$ so that ${\hat U}^T {\hat U} = I_{2 \times 2}$ and ${\hat V}^T {\hat V}=I_{2 \times 2}$. Then, using \eqref{eq:uv},
\begin{align*}
A &= UV^T =  \left(n \sum {\alpha_i}^2\right)^\frac{1}{2} ~ \hat U \begin{bmatrix} \sigma & 0 \\ 0 &  \sigma \end{bmatrix} \hat V^T.
\end{align*}

 This thin SVD decomposition of $A$ implies that the thin SVD decomposition of $E$ is of the form \eqref{eq:E_svd}. Using Lemma \ref{lemma:SVDtoSPEC}
we obtain that ${\hat V} = \sqrt{0.5}(X+Y)$ where $X$ and $Y$ include the eigenvectors of $E,$ and the sub-blocks of $\hat{V}$ are orthogonal, since $\hat{V}_i = \frac{1}{\sqrt{n}}R_i^T Q$ satisfies ${\hat V}_i^T {\hat V}_i=\frac{1}{n} I_{2 \times 2}$, $i \in [n]$, implying that the second condition is satisfied.

$\Leftarrow$:
We show that if $E \in {\cal E}$ %, an $n$-view essential matrix
 satisfies conditions      1-2,  $E$ is consistent. Following condition 1, $E$ is of rank 4 with spectral decomposition of the form as in Lemma \ref{lemma:SVDtoSPEC}
and therefore, due to this lemma, its SVD takes the form $E=\lambda\hat{U}\hat{V}^T+\lambda\hat{V}\hat{U}^T$, where $\hat U,\hat V \in \Real^{3n \times 2}$ are given by ${\hat  U} = \sqrt{0.5} (X-Y)$ and  ${\hat V} = \sqrt{0.5} (X+Y)$. Moreover, due to condition (2), every $3 \times 2$ block  ${\hat V}_i$ satisfies ${\hat V}_i^T {\hat V}_i = \frac{1}{n} I_{2 \times 2}$.

Our next aim is to use  ${\hat U}$ and ${\hat V}$ to define $U, V \in \Real^{3n \times 3}$ in a form that implies the consistency of $E$. For each $i \in [n]$ we construct a $3 \times 3$ block ${\tilde V}_i$ by adding a third column to ${\hat V}_i$ that is set to be the cross product of its two columns, scaled by $\sqrt{n}$. The sign of this vector is further selected such that $\det({\tilde V}_i) > 0$. Consequently, $R_i=\sqrt{n} {\tilde V}_i$ is a rotation matrix, and we let
\begin{equation}\label{eq:defV}
V = \begin{bmatrix} R_1, ..., R_n \end{bmatrix}^T.
\end{equation}
Additionally, we pad  ${\hat U}$ with a third column of zeros, and then scale the obtained matrix by $\lambda/\sqrt{n}$, yielding $U \in \Real^{3n \times 3}$. It can be readily verified that  $E = UV^T + VU^T$.

Next, since $E$ is an $n$-view essential matrix, $E_{ii} = 0,$ implying that $U_i V_i^T$ is skew symmetric. This means that there are two possible options: either $\rank(U_i V_i^T) = 2$ or $\rank(U_i V_i^T) = 0$. Since $\rank(V_i)=3$ this respectively implies that either $\rank(U_i)=2$ or $U_i = 0$. By Lemma \ref{lemma:skew}, 
when $\rank(U_i)=2$ and $\rank(V_i)=3$ then
\begin{equation}  \label{eq:Ti}
T_i=V_i^T U_i
\end{equation}
is skew-symmetric. Moreover, since the third column of $U_{i}$ is identically zero, then $T_i = \alpha_i T$ for some $\alpha_i \in \Real$ and $T = \begin{bmatrix} 0 & -1 & 0 \\ 1 & 0 & 0 \\ 0 & 0 & 0\end{bmatrix}$. This also includes the case that $U_i=0$ by setting $\alpha_i=0$.
Combining \eqref{eq:defV} and \eqref{eq:Ti}, $U_i = V_i^{-T} T_i = \alpha_i R_i^T T$, implying that $U=\begin{bmatrix} \alpha_1  R_1^T \\ \vdots \\ \alpha_n R_n^T \end{bmatrix} T$.  Finally, $E=UV^T+VU^T$ and the forms of $U$ and $V$ imply that $E$ is consistent and realized by cameras with collinear centers.
\end{proof}

\section{Proof of Theorem~\ref{thm:fundamental}}  \label{sec:fundamental_proof}

\begin{proof}
$\Rightarrow:$ Assume $F \in {\cal F}$ is consistent and realized by cameras with collinear centers. Then, by definition and due to collinearity $F$ can be formulated as $F=A+A^T $ where $A=UV^{T}$and $U,V \in \Real^{3n \times 3}$,
\begin{align*}
U =\left[\begin{array}{c}
\alpha_1 V_{1} \\
\vdots\\
\alpha_n V_{n}
\end{array}\right]T,
~~ & ~~
V =\left[\begin{array}{ccc}
V_{1}\\
\vdots\\
V_{n}
\end{array}\right],
\end{align*}
where $V_i \in \Real^{3 \times 3}$ are full rank, $T = [{\bf t}]_{\times}$ and ${\bf t} \in \Real^3$.
%${\bf t}_i = \alpha_i {\bf t}$, such that $ \sum_{i=1}^n {\bf t}_i=\sum_{i=1}^n \alpha_i {\bf t} = 0$.

Next, we show that $\rank(F)=4$.
Using QR decomposition we can write $V_i = K_i^{-T} R_i^T$, where $K_i$ is invertible and upper triangular and $R_i \in SO(3)$. This decomposition is unique due to the full rank of $V_i$. We can therefore write $F=K^T E K$, where the $3n \times 3n$ matrix $K$ is block diagonal with $3 \times 3$ blocks formed by $\{K_i^{-1}\}_{i=1}^n$ and thus  has full \rank. As a consequence $\rank(F) = \rank(E)$. By construction $E$ has the same form as in Thm.~$\ref{thm:essential}$, and therefore $\rank(E) = 4$, implying also that $\rank(F) = 4$.

Next, since $\rank(F)=4$ and $\rank(U)=2$ we have that $\rank(A)=2$ and we can express its full-rank decomposition as $A={\tilde U}{\tilde V}^T$, where ${\tilde U, \tilde V} \in \Real^{3n \times 2}$ and $\rank({\tilde U})=\rank({\tilde V})=2$. Using Lemma \ref{lemma:eigenvalues} 
$F$ has two positive and two negative eigenvalues. This concludes the proof of the first condition.

The second condition is justified as follows. Due to collinearity, $\exists {\bf e} \in \Real^3, {\bf e} \neq 0$, s.t.\ $F_{ji} {\bf e} = 0$ for some camera  $\exists i \in [n]$ and for all other cameras $j \in [n]$, implying that $F_i^T {\bf e}=0$. Consequently, for all $i \in [n]$, $\rank(F_i) \leq 2$, implying that $\rank(F_i) = 2$, since otherwise if  $\rank(F_i) < 2$ it contradicts the assumption that $\rank(F_{ji})=2$, for all $j$.

$\Leftarrow$: Let $F \in {\cal F}$ then if $F$ satisfies conditions 1 and 2. We prove that it is consistent and realized by cameras with collinear centers. Condition 1 and Lemmas \ref{lemma:eigenvalues} 
imply that $F$ can be decomposed into $F = {\hat U}{\hat V}^T + {\hat V}{\hat U}^T,$ where ${\hat U}, {\hat V} \in \Real^{3n \times 2}$ and $\rank({\hat U})=\rank({\hat V})=2$ and all $3 \times 2$ blocks are full rank. 

We next use ${\hat U}$ and ${\hat V}$ to define $U, V \in \Real^{3n \times 3}$ in a form that implies the consistency of  $F$ as follows. $U= \begin{bmatrix} {\hat U}, 0 \end{bmatrix}$ and for each $i \in [n]$ we construct a $3 \times 3$ block $V_i$ by appending a third column to ${\hat V}_i$, setting it to be the cross product of its two columns. Since $F_{ii} = 0$ for all $i$, then ${\hat U}_i {\hat V}_i^T = U_i V_i^T$ is skew-symmetric, and by construction  $\rank(U_i) =2$ and  $\rank(V_i)=3$. Therefore  by Lemma \ref{lemma:skew}, 
$T_i=V_i^{-1} U_i$ is also skew-symmetric. Therefore, $T_i = [{\bf t}_i]_{\times}$, for some ${\bf t}_i \in \Real^3$, implying that  $F_{ij} = V_i(T_i-T_j)V_j^T$ and therefore that $F$ is an $n$-view fundamental matrix.

Finally, collinearity follows from the second condition, which implies that $\exists {\bf e} \neq 0$ s.t. $F_i^T {\bf e} = 0$. This implies that $F_{ji} {\bf e} = 0$ for all $j \in [n]$, asserting that all $n$\ cameras are collinear.

\end{proof}

\section{Supporting lemmas}
Below are a few supporting lemmas which Thms.~\ref{thm:essential} and ~\ref{thm:fundamental} rely on.  

\begin{lemma}\label{lemma:SVDtoSPEC} Let $E\in \mathbb{S}^{3n}$ of rank 4, and $\Sigma \in \Real^{2 \times 2} $, a diagonal matrix, with positive elements on the diagonal. Let $X, Y, U, V \in \Real^{3n \times 2}$, and we define the mapping $(X,Y) \leftrightarrow (U,V):$ $X = \sqrt{0.5}({\hat U}+{\hat V})$, $Y = \sqrt{0.5}({\hat V}-{\hat U})$, ${\hat U} = \sqrt{0.5}(X-Y), {\hat V} = \sqrt{0.5}(X+Y)$.

Then, the (thin) SVD of $E$ is of the form $$E=\left[ {\hat{U}},{\hat{V}}\right]\left(\begin{array}{cc}
\Sigma\\
 & \Sigma
\end{array}\right)\left[\begin{array}{c}
 {\hat{V}}^{T}\\
 {\hat{U}}^{T}
\end{array}\right]$$   if and only if the (thin) spectral decomposition of $E$ is
of the form

$$E=[X, Y]\left(\begin{array}{cc}
\Sigma\\
 & -\Sigma
\end{array}\right)\left[\begin{array}{c}
 X^{T}\\
 Y^{T}
\end{array}\right].$$
\end{lemma}
\begin{proof}
The proof is similar to  Lemma 6 presented in the supplementary material of \cite{kasten2019algebraic}. 
\end{proof}

\begin{lemma}\label{lemma:eigenvalues}
Let $F \in \mathbb{S}^{3n}$ be a matrix of rank 4. Then, the following three conditions are equivalent.
\begin{enumerate}
\item[(i)] $F$ has exactly 2 positive and 2 negative eigenvalues.
\item[(ii)] $F=XX^T-YY^T$ with $X,Y \in \Real^{3n \times 2}$ and $\rank(X)=\rank(Y)=2$.
\item[(iii)] $F=UV^T+VU^T$ with $U,V \in \Real^{3n \times 2}$ and $\rank(U)=\rank(V)=2$.
\end{enumerate}
\end{lemma}
\begin{proof}
The proof is similar to the proof given in \cite{kasten2018gpsfm}, Lemma 1.
\end{proof}

\begin{lemma}\label{lemma:skew} Let $A,B\in \Real^{3\times3}$ with $\rank(A)=2$, $\rank(B)=3$ and $AB^{T}$ is skew symmetric, then
$T=B^{-1}A$ is skew symmetric.
\end{lemma}
\begin{proof}
See Lemma 4 in the supplementary material of \cite{kasten2018gpsfm}.
\end{proof}

\begin{lemma}\label{skew_symmetric_2_}
Let $A,B\in \mathbb{R}^{2\times 2}$ s.t $AB^T \ne 0$ and skew-symmetric. Then it follows that $A=BT_y$ where $T_y$ is $2\times 2$ skew symmetric.
\end{lemma}

\begin{proof}
First we note that $AB^T=T_x$, where
$$T_x=\begin{bmatrix}0 & x \\-x & 0\end{bmatrix}$$
for some $0 \ne x \in \Real$ and $\rank(A)=\rank(B)=2$. Therefore,
$$A = T_x B^{-T} = B B^{-1} T_x B^{-T}$$
Denote by $T_y = B^{-1}T_xB^{-T}$. Clearly $T_y$ is skew symmetric, since
\[ T_y^T = (B^{-1}T_xB^{-T})^T =  B^{-1}T_x^TB^{-T} = - B^{-1}T_xB^{-T}\]
where the rightmost equality is due to the skew symmetry of $T_x$. 

%$$\begin{bmatrix} a& b \\c & d\end{bmatrix}\begin{bmatrix}0 & x \\-x &  0\end{bmatrix} \begin{bmatrix}a & c \\ b & d\end{bmatrix}=\begin{bmatrix}-bx& ax \\-dx & cx\end{bmatrix}\begin{bmatrix}a & c \\ b & d\end{bmatrix}=$$$$\begin{bmatrix}-bxa+abx & -cbx+adx \\-adx+cxb &-dxc+dxc\end{bmatrix}=\begin{bmatrix}0 & -(cbx-adx) \\cxb-adx &0\end{bmatrix}=T_y$$

\end{proof}

\begin{lemma}\label{skew_symmetric}
Let $A,B\in \mathbb{R}^{3\times 2}$ with $rank(A)=rank(B)=2$ and $AB^T$ is skew symmetric. Then $B=AT$ for some skew symmetric matrix $T\in \mathbb{R}^{2\times 2}$.
\end{lemma}

\begin{proof}
$$AB^T=\begin{bmatrix}
A_u \\ \mathbf{a}^T
\end{bmatrix}\begin{bmatrix}
B_u^T & \mathbf{b}
\end{bmatrix}=\begin{bmatrix} A_u B_u^T  & A_u \mathbf{b} \\ \mathbf{a}^T B_u^T & \mathbf{a}^T \mathbf{b}\end{bmatrix}$$ Where $A_u,B_u\in\mathbb{R}^{2\times 2}$ and $\av,\bv \in \Real^2$.
As a result $$\mathbf{a}^T\mathbf{b}=0, ~A_u\mathbf{b}=-B_u\mathbf{a}$$
%and we denote by $T_1=A_uB_u^T$.

Assuming first that $A_uB_u^T\neq0$ we get (using Lemma \ref{skew_symmetric_2_}) that $$A_u=B_uT$$
with $T \in \Real^{2 \times 2}$ skew symmetric. Consequently,
$$AB^T=\begin{bmatrix}B_uTB_u^T &B_uT\mathbf{b} \\ -\mathbf{b}^T T^TB_u^T & 0 \end{bmatrix}$$
Since $\bv^T T \bv=0$ we can write 
$$AB^T=\begin{bmatrix}B_uTB_u^T &B_uT\mathbf{b} \\ \mathbf{b}^TTB_u^T & \mathbf{b}^TT\mathbf{b}\end{bmatrix} = %\begin{bmatrix}B_{u}T_0 \\ \mathbf{b}^TT_0
%\end{bmatrix} \begin{bmatrix}B_{u}^T & \mathbf{b}
%\end{bmatrix}$$
\begin{bmatrix}B_{u} \\\mathbf{b}^T
\end{bmatrix}T \begin{bmatrix}B_{u}^T & \mathbf{b}
\end{bmatrix}
= BTB^T$$
Since $\rank(B)=2$, multiplying by $B(B^TB)^{-1}$ yields,
$$AB^TB(B^TB)^{-1}=BT_0B^TB(B^TB)^{-1}$$
and consequently,
$$A=BT$$
A similar construction can be made in case $A_uB_u^T=0$.

\begin{comment}
\yk{I did the next part only for verification - I think that we can remove it from the final proof and just say that we can decompose the lower part when $A_uB_u^T=0$}
If $A_uB_u^T=0$, since the rank of $A,B$ is 2 then   $rank(A_u)>0,rank(B_u)>0$. 
As a result,   $A_u[b_1,b_2]=0\Rightarrow [A_ub_1,A_ub_2]=0$ which means that $A_u$ is rank deficient so its rank is one and as a result the rank of B is 1. As a result $A=\begin{bmatrix} a_1^T\\\alpha a_1^T \\a_3^T
\end{bmatrix}$ $B=\begin{bmatrix} b_1^T\\\beta b_1^T \\b_3^T
\end{bmatrix}$.

As a result both $a_3, b_3$ are not a scaled versions of $a_1,b_1$ respectively otherwise the ranks of both $A,B$ is one. Then defining $$AB^T=\begin{bmatrix}a_1^T \\A_b\end{bmatrix}\begin{bmatrix}b_1 & B_b^T\end{bmatrix}$$
We get:
$$AB^T = \begin{bmatrix}a_1^Tb_1 & a_1^T B_b^T \\A_b b_1 & A_bB_b^T\end{bmatrix}$$
Since $A_b,B_b$ are full rank we get: 
$$A_bb_1=-B_ba_1,A_b=B_bT_0,a_1^Tb_1=0$$
And we get:
$$AB^T= \begin{bmatrix}0 & -b_1^TT_0^TB_b^T \\B_bT_0b_1 & B_bT_0B_b^T\end{bmatrix}=\begin{bmatrix}b_1^TT_0b_1 & b_1^TT_0B_b^T \\B_bT_0b_1 & B_bT_0B_b^T\end{bmatrix}$$
$$AB^T=\begin{bmatrix}b_1T_0\\ B_bT_0\end{bmatrix}\begin{bmatrix}b_1 & B_b^T\end{bmatrix}=\begin{bmatrix}b_1\\ B_b\end{bmatrix}T_0\begin{bmatrix}b_1 & B_b^T\end{bmatrix}=BT_0B^T$$
$$A=BT_0$$
\end{comment}

\end{proof}

\begin{lemma}\label{orthogonal}
Suppose that $T\in \mathbb{R}^{2\times 2}$ is a non-zero skew symmetric matrix, and $\uu, \vv \in \Real^2$ are orthonormal. Then, (1) it holds that for any vector $\xx \in \Real^2$, $T\xx$ is perpendicular to $\xx$ and  (2) if $\xx$ is expressed in terms of the orthonormal basis as $\xx = \alpha \uu + \beta \vv$ then $T\xx$ is parallel to $\beta \uu - \alpha \vv$.
\end{lemma}

\begin{proof}
Since $T \in \Real^{2 \times 2}$ is skew-symmetric, then for all $\xx\in \mathbb{R}^2$ it holds that $\xx^T T \xx=0$, implying that $T \xx$ is perpendicular to $\xx$. If $\xx=\alpha \uu + \beta \vv$, where $\uu, \vv \in \Real^2$ are orthonormal then clearly $(\beta \uu - \alpha \vv)^T\xx=0$. Due to the uniqueness of orthogonality in $\Real^2$ (up to scale), we obtain that $T\xx$ is parallel to $\beta \uu - \alpha \vv$.
\end{proof} 

\begin{lemma}\label{lemma:rep2}
Let $F \in \Symmetric^{3n}$ such that (1) $F=UV^T+VU^T$ with $U, V \in \Real^{3n \times 2}$ and (2) $\forall i\in[n]$, $F_{ii}=U_iV_i^T+V_iU_i^T=0$. Suppose that $\exists i$ such that  $\rank(U_i)=\rank(V_i)=1$ and $U_i V_i^T =0$, and $\exists j\neq i$ $\rank(U_j)=\rank(V_j)=2$, then $\rank(F_{ij}) = U_iV_j^T + V_i U_j^T \leq 1$.
\end{lemma}

\begin{proof}
W.l.o.g. assume $\rank(U_1)=\rank(V_1)=1$ and  $\rank(U_2)=\rank(V_2)=2$. Since $V_2 U_2^T$ is skew-symmetric and $\rank(V_2)=\rank(U_2)=2$, it follows from Lemma \ref{skew_symmetric}, that  $U_2$ can be expressed as 
\begin{equation}\label{eq:UVT} U_2=V_2 T \end{equation} 
where $T \in \Real^{2 \times 2}$
%$T=\begin{bmatrix} 0 & a \\ -a & 0 \end{bmatrix}$ is $2 \times 2$ 
is a skew-symmetric matrix.

Since $\rank(U_i) = \rank(V_i) =1$, they can be expressed as 
\begin{equation}\label{eq:UV_rank1} 
U_1 = \begin{bmatrix} \alpha_0 \uu^T \\ \alpha_1 \uu^T \\ \alpha_2 \uu^T \end{bmatrix},~~~V_1 = \begin{bmatrix} \beta_0 \vv^T \\ \beta_1 \vv^T \\ \beta_2 \vv^T \end{bmatrix}\end{equation}
where $\uu, \vv \in \Real^{2}$ with $\norm{\uu}=\norm{\vv}=1$, $\{\alpha_i\}$ and $\{\beta_i\}$ are scalars. Moreover, since we assume that $U_1 V_1^T = 0$, it follows that $\uu^T \vv=0$. Therefore, $\{\uu, \vv\}$ form an orthonormal basis in $\Real^{2}$ and thus $\exists \{\gamma_i\}$ and $\{\delta_i\}$ such that 
\begin{equation}\label{eq:V2} V_2^T = \begin{bmatrix}\gamma_1 \uu + \delta_1 \vv, & \gamma_2 \uu + \delta_2 \vv, & \gamma_3 \uu + \delta_3 \vv \end{bmatrix}_{2 \times 3} \end{equation}
and by \eqref{eq:UVT} and using Lemma \ref{orthogonal} also 
\begin{equation}\label{eq:U2} U_2^T=-T V_2^T  =  a \begin{bmatrix} \gamma_1 \vv - \delta_1 \uu, & \gamma_2 \vv - \delta_2 \uu, & \gamma_3 \vv - \delta_3 \uu \end{bmatrix}_{2 \times 3} \end{equation}
for some scalar $a$.  

Using \eqref{eq:UV_rank1}  and \eqref{eq:V2}, we obtain
$$U_1 V_2^T = \begin{bmatrix} \alpha_0 \\ \alpha_1 \\ \alpha_2 \end{bmatrix} \begin{bmatrix} \gamma_1 & \gamma_2 & \gamma_3\end{bmatrix}.$$ 
In a similar way, using \eqref{eq:UV_rank1} and \eqref{eq:U2} we obtain 
$$V_1 U_2^T = a \begin{bmatrix} \beta_0 \\ \beta_1 \\ \beta_2  \end{bmatrix} \begin{bmatrix}  \gamma_1 & \gamma_2 & \gamma_3 \end{bmatrix}.$$
Finally, 
$$F_{12} = U_1 V_2^T + V_1 U_2^T = \left(\begin{bmatrix} \alpha_1 \\ \alpha_2 \\ \alpha_3 \end{bmatrix} + a \begin{bmatrix} \beta_0 \\ \beta_1 \\ \beta_2 \end{bmatrix} \right) \begin{bmatrix} \gamma_1 & \gamma_2 & \gamma_3 \end{bmatrix}$$
implying that $\rank(F_{12}) \leq 1$

\end{proof}

\begin{lemma}\label{lemma:representation}
Let $F \in {\cal F}$, a n-view  fundamental matrix, with $\rank(F) = 4$ such that (1) $F=U V^T + V U^T$ with $U,V \in \Real^{3n \times 2}$, $\rank(U) = \rank(V) = 2$,  (2) $\rank(F_i) = 2$. Then, $F$ can be written also as $F=\hat U \hat V^T + \hat V \hat U^T$ with $\hat U, \hat V \in \Real^{3n \times 2}$ such that for all $i \in [n]$, $\hat U_i \hat V_i^T$ is skew-symmetric and $\rank(\hat U_i \hat V_i^T) = 2$.
\end{lemma}

\begin{proof}
Since  $F_{ii}=0$ for all $i \in [n]$, $U_i V_i^T + V_i U_i^T =0$, and hence $U_i V_i^T$ is skew-symmetric. Therefore, either $\rank(U_i V_i^T)=2$ or $U_i V_i^T=0$. We thus need to show that there exists a decomposition of $F$ such that $\hat U_i \hat V_i^T \ne 0$ for all $i \in [n]$.

Suppose that $U_i V_i^T=0$ for some $i \in [n]$. First, we note that both cases, that  $\rank(U_i)=\rank(V_i)=2$ and  $\rank(U_i)=1$, $\rank(V_i)=2$, are not feasible. The former case is infeasible because the null space of $U_i$ only includes the zero vector. The latter case is infeasible because the null space of $U_i$ is of dimension 1, and the space spanned by the rows of $V_i$ is of dimension 2. Consequently, it remains to analyze the case that either (i) $\rank(U_i) = \rank(V_i) = 1$, and (ii) $U_i =0$ (or $V_i =0$)  for some $i \in [n]$. %We next analyze the case that $\rank(U_i)=\rank(V_i)=1$ such that $U_i V_i^T =0$ and 
We further prove that it is possible to generate a valid decomposition, i.e., for all $i \in [n]$, $\hat U_i \hat V_i^T$ is skew-symmetric and $\rank(\hat U_i \hat V_i^T) = 2$. 
%The latter case is discussed subsequently.

\begin{comment}
\section{Preliminary proof for the case of $U_iV_i^T=0$}
\subsection{Proof that $rank1,rank2$ is not possible}
\ag{would we like to write it as a lemma? Maybe just to mention that this case is not possible?}

Let $U_i,V_i\in \mathbb{R}^{3\times 2}$ such that $rank(U_i)=1,rank(V_i)=2$

Then it follows that: $$U_i=\begin{bmatrix}u^T \\bu^T \\cu^T \end{bmatrix},V_i=\begin{bmatrix}v_1^T \\v_2^T \\ v_3^T\end{bmatrix}$$
Where $u,v_1,v_2,v_3\in \mathbb{R}^{2\times1}$

As a result assume that $$U_iV_i^T=0\Rightarrow \begin{bmatrix}u^T \\bu^T \\cu^T \end{bmatrix} \begin{bmatrix}v_1 & v_2 & v_3\end{bmatrix}=\begin{bmatrix}u^Tv_1 &u^Tv_2 & u^Tv_3\\ bu^Tv_1 &bu^Tv_2 & bu^Tv_3\\cu^Tv_1 &cu^Tv_2 & cu^Tv_3 \end{bmatrix}=0$$

So we have $u^Tv_1=0,u^Tv_2=0,u^Tv_3=0$

However since $u^T \in \mathbb{R}^{1\times2}$ with rank 1,we get that its null space is of dim 1. As a result $\exists \alpha,\beta $ such that: $v_1=\alpha v_2= \beta v_3$ 

($A\in\mathbb{R}^{p\times q}\Rightarrow q=rank(A)+dimN_A$
in our case $A=u^T\in\mathbb{R}^{1 \times 2}\Rightarrow q=2,rank(A)=1 \Rightarrow 2=1+dimN_A\Rightarrow dimN_A=1$)

This is a contradiction to the fact that $rank(V_i)=2$ 
%$u\in \mathbb{R}^{2\times 1} \Rightarrow q=1 \Rightarrow 1=1+dimN_A\Rightarrow dimN_A=0$
\end{comment}

Following the assumptions, $\rank(F)=4$ and $F = UV^T + VU^T$, such that $\rank(U) =\rank(V) = 2$, and using Lemma \ref{lemma:eigenvalues}, it holds that $F= XX^T-YY^T$ where $X,Y\in \mathbb{R}^{3n\times2}$, $\rank(X) = \rank(Y) =2$ and we denote  $X = [\xx^1, \xx^2]$ and $Y = [\yy^1, \yy^2]$.  The entries of the vectors are enumerated by subscript $i$.

Based on the $X, Y$ decomposition, we will construct $\hat{U}, \hat{V} \in \Real^{3n \times 2}$, such that $F = \hat{U} \hat{V}^T + \hat{V} \hat{U}^T$, where for any $i \in [n]$, it holds that $\hat{U}_i \hat{V}_i^T$ is skew-symmetric with $\rank(\hat{U}_i \hat{V}_i^T)=2$. 
More precisely, we use either 
\begin{equation}\label{eq:rep1}\hat{U}=\frac{1}{\sqrt{2}}[\xx^1+\yy^1,\xx^2-\yy^2],~~~\hat{V}=\frac{1}{\sqrt{2}}[\xx^1-\yy^1,\xx^2+\yy^2]\end{equation}
or 
\begin{equation}\label{eq:rep2}\hat{U}=\frac{1}{\sqrt{2}}[\xx^1+\yy^1,\xx^2+\yy^2],~~~\hat{V}=\frac{1}{\sqrt{2}}[\xx^1-\yy^1,\xx^2-\yy^2]\end{equation}
to construct a decomposition for $F$ that satisfies the conditions of the lemma. Both decompositions satisfy that $F = XX^T - YY^T =\xx^1 \otimes \xx^1 + \xx^2 \otimes \xx^2 - \yy^1 \otimes \yy^1 - \yy^2 \otimes \yy^2 = \hat{U} \hat{V}^T + \hat{V} \hat{U}^T$, where we use the symbol $\otimes$ to the denote the outer product. 
%Next, we show that one of the decompositions satisfies that for any $i \in [n]$, $\rank(\hat{U}_i) = \rank(\hat{V}_i) = 2$.

\textbf{Special case: $\exists i$ such that $U_i=0$ or $V_i=0$.}
Assuming that for one of the decompositions, \eqref{eq:rep1} and \eqref{eq:rep2}, there exists $i$ where w.l.o.g. $U_i=0$. Then we note that $\rank(V_i)=2$ since otherwise $\rank(F_i)<2$.  Moreover, $\forall j\neq i$ $\rank(U_j)=2$ since otherwise $\rank(F_{ij})<2$, and therefore for all $j\neq i$ $\rank(V_j)\in \{0,2\}$. Consequently, we can use Lemma \ref{lemma:Qrep} to construct the desired decomposition. Hence from now on we discard the case that $\exists i$ such that $U_i=0$ or $V_i=0$.

\textbf{First step: for any $i \in [n]$ in one of the decompositions, \eqref{eq:rep1} or \eqref{eq:rep2}, it holds that $\rank(\hat{U}_i) = \rank(\hat{V}_i) = 2$.}
%\textbf{First step: the case that $\rank(\hat{U}_i) = \rank(\hat{V}_i)  = 1$ for some $i \in [n]$ in both decompositions, \eqref{eq:rep1} and \eqref{eq:rep2}, is not feasible. }
W.l.o.g. we prove this for $U_1,V_1$. Assume by contradiction that in both decompositions  $\rank(\hat {U}_1)=\rank(\hat{V}_1)=1$. 
First, we note that in both decompositions, the situation that one of the columns of $\hat{U}_1$ or $\hat{V}_1$ is identically zero, is not possible. W.l.o.g., assume that the first column of $\hat{U}_1$ in the first representation, $\xx_{(1:3)}^1 + \yy_{(1:3)}^1$, is zero. Then due to the rank 1 assumption and the relation $\hat{U}_1 \hat{V}_1^T =0$, this implies that $\xx_{(1:3)}^2 + \yy_{(1:3)}^2$ is zero, yielding $U_1$ which is identically zero in the second representation, contradicting the $\rank 1$ assumption.

Now, due to the $\rank 1$ assumption, we obtain that 
the following vectors (none of them is zero) are parallel
$$\xx_{(1:3)}^1 + \yy_{(1:3)}^1,~~~ \xx_{(1:3)}^1 - \yy_{(1:3)}^1,~~~ \xx_{(1:3)}^2 + \yy_{(1:3)}^2,~~~ \xx_{(1:3)}^2 - \yy_{(1:3)}^2 \Rightarrow$$

$\exists \uu \neq 0 \in \Real^{3 \times 1}$ and scalars $\{\gamma_i \neq 0 \}_{i=1}^4$ such that 
 \begin{equation}\label{UV:eq}
U_1=[\gamma_1 \uu,\gamma_2  \uu] ~~V_1=[\gamma_3 \uu,\gamma_4 \uu]
\end{equation}
Therefore, 
\begin{align*}F_1 &= U_1 V^T + V_1 U^T \\
& = \begin{bmatrix} \gamma_1 \uu & \gamma_2 \uu \end{bmatrix} V^T + \begin{bmatrix} \gamma_3 \uu  & \gamma_4 \uu \end{bmatrix} U^T \\
& = \begin{bmatrix} \gamma_1 \uu & \gamma_2 \uu & \gamma_3 \uu & \gamma_4 \uu  \end{bmatrix} \begin{bmatrix} V^T \\ U^T \end{bmatrix}
\end{align*}
yielding $\rank(F_1) \leq 1$, contradicting the assumption that $\rank(F_i) = 2$.

\begin{comment}
\ag{Treatment to some minor case (which I think we don't need to include): In our analysis, it holds that for $i=1,2$ $X_i^1\pm Y_i^1 \neq \vec 0$. The reason for this is because if for example $X_1^1+Y_1^1=\vec 0$ then in order to satisfy $U_1V_1^T=0$ using representation (1) it must hold that $X_1-Y_1=\vec 0$. Then by applying the same argument, since $X_1-Y_1=\vec 0$ using representation (2) it must hold that $X_2+Y_2=\vec 0$, since we began with $X_1^1+Y_1^1=\vec 0$ it implies that in representation (2) $\hat{U}_1=0$ but we treat in the above only the case where $\rank (\hat U_1)=1$ so it implies that $i=1,2$ $X_i^1\pm Y_i^1 \neq \vec 0$  }\\
\ag{When reading the parallelizem relation note that all of the the vector below are not zero, if one of them is zero then (together with $U_1V_1^T=0$) it collapses to the case that either (1) or (2) give parametrization with one block which is zero (which we solve in the next sub section) }
\ag{This part ignores the case that one of the blocks is zero. We  prove it in 3.3. Note that if non of the blocks is zero, then in the two options (1),(2) all the columns are different than zero}

\end{comment}

{\bf Second step: Each decomposition satisfies $\forall i~~~ \rank(\hat{U}_i) = \rank(\hat{V}_i) =1$ or  $\forall i~~~\rank(\hat{U}_i) = \rank(\hat{V}_i) =2$.}
Following the first step, we get that for any $i \in [n]$, it holds that $\rank(\hat{U}_i) = \rank(\hat{V}_i) = 2$ in one of the decompositions. Select $j \in [n]$ and assume w.l.o.g. that $\rank(\hat{U}_j) = \rank(\hat{V}_j) = 2$ in the first decomposition.  Now, for each $i$, we know that $\hat{U}_i \hat{V}_i^T$ is a skew symmetric matrix. Therefore, for any $i\neq j$ $\rank(\hat{U}_i \hat{V}_i^T)$ is either 2 or 0. If the rank is 2, then immediately $\rank(\hat{U}_i) = \rank(\hat{V}_i)=2$. If the rank is 0, i.e., $\hat{U}_i \hat{V}_i^T =0$, then $\rank(\hat{U}_i) = \rank(\hat{V}_i)=1$ and this is contradicted using Lemma \ref{lemma:rep2}, and therefore $\rank(\hat{U}_i) = \rank(\hat{V}_i)=2$.

These proof shows that there is a decomposition such that for any $i$, it holds that $\rank(\hat{U}_i) = \rank(\hat{V}_i)=2$.

\end{proof}

\begin{lemma} \label{lemma:Q_2_2}
Let $F \in {\cal F}$,
$\rank(F)=4$ and $F=UV^T+VU^T$ where $U,V\in \mathbb{R}^{3n\times 2}$. Let $Q=\begin{bmatrix} Q_{11} & Q_{12} \\ Q_{21} & Q_{22}\end{bmatrix}\in \mathbb{R}^{4\times 4} $ such that it is subject to the following constraints:
\begin{enumerate}
    \item $Q_{12}Q_{11}^T+Q_{11}Q_{12}^T=0$, i.e., $Q_{12}Q_{11}^T$ is skew symmetric (3 constraints)
    \item $Q_{22}Q_{21}^T+Q_{21}Q_{22}^T=0$, i.e., $Q_{22}Q_{21}^T$ is skew symmetric (3 constraints)
    \item $Q_{12}Q_{21}^T+Q_{11}Q_{22}^T=I$ (4 constraints)
\end{enumerate}
It then follows that $F=\tilde{U}\tilde{V}^T+\tilde{V}\tilde{U}^T$ where $[\tilde{U} \  \tilde{V}]=[U \ V]Q$.

This Lemma is adapted from \cite{sengupta2017new} that proved a similar result for the case that $\rank(F)=6$.
\end{lemma}

\begin{proof}
Denote by $J=\begin{bmatrix}0 &I  \\I & 0\end{bmatrix}\in \mathbb{R}^{4\times 4}$, then,
$$QJQ^T=
%\begin{bmatrix} Q_{11} & Q_{12} \\ Q_{21} & Q_{22}\end{bmatrix} J \begin{bmatrix} Q_{11} & Q_{12} \\ Q_{21} & Q_{22}\end{bmatrix}^T=
\begin{bmatrix} Q_{11} & Q_{12} \\ Q_{21} & Q_{22}\end{bmatrix} \begin{bmatrix}0 &I  \\I & 0\end{bmatrix}\begin{bmatrix} Q_{11} & Q_{12} \\ Q_{21} & Q_{22}\end{bmatrix}^T=
\begin{bmatrix}Q_{12} & Q_{11} \\ Q_{22} &Q_{21}\end{bmatrix}\begin{bmatrix} Q_{11}^T & Q_{21}^T \\ Q_{12}^T & Q_{22}^T\end{bmatrix}=$$
$$\begin{bmatrix}Q_{12}Q_{11}^T+Q_{11}Q_{12}^T & Q_{12}Q_{21}^T+Q_{11}Q_{22}^T \\ Q_{22}Q_{11}^T+Q_{21}Q_{12}^T & Q_{22}Q_{21}^T+Q_{21}Q_{22}^T\end{bmatrix}=\begin{bmatrix}0 &I  \\I & 0\end{bmatrix}$$
where the rightmost equality is obtained by using the constraints on $Q$. Let $\tilde{U},\tilde{V}\in\mathbb{R}^{3n\times 2}$ defined as $[\tilde{U} \  \tilde{V}]=[U \ V]Q$. It follows that: $$F=UV^T+VU^T=[U \ V]J[U \ V]^T=[U \ V]QJQ^T[U \ V]^T=[\tilde{U} \ \tilde{V}]J[\tilde{U} \ \tilde{V}]^T=\tilde{U}\tilde{V}^T+\tilde{V}\tilde{U}^T$$
\end{proof}

\begin{lemma}\label{lemma:Qrep}
Let $F \in {\cal F}$,
%Let $F \in \mathbb{S}^{3n}$  be an $n$-view fundamental matrix. 
for which the following conditions hold: 
\begin{enumerate}
\item $\rank(F)=4$ and $F=UV^T+VU^T$, where $U,V\in \mathbb{R}^{3n\times 2}$
\item $\rank(F_i) = 2,$ where $F_i$ denotes the $i^{th}$ block-row of $F$,  $i \in [n]$.
\item For all $i\in [n]$, $\rank(U_i),\rank(V_i) \in \{0,2\}$, where $U_i$ and $V_i$ respectively denote the $3\times 2$ blocks of $U,V$, and there exists at least one block of rank 0.
\end{enumerate}
Then there exists a new decomposition $F=\tilde{U}\tilde{V}^T+\tilde{V}\tilde{U}^T$ such that all the $3\times 2$ blocks of $\tilde{U},\tilde{V}$ are of rank 2.
\end{lemma}
\begin{proof}
Following Lemma \ref{lemma:Q_2_2}, let $\alpha\neq 0 $ we set $Q$ as follows: 
$$Q_{11}=\frac{1}{\alpha}\begin{bmatrix}1 & 0 \\  0 & 1 \end{bmatrix},Q_{12}=\begin{bmatrix}0 & -1 \\  1 & 0 \end{bmatrix},Q_{21}=\begin{bmatrix}0 & -2 \\  2 & 0 \end{bmatrix},Q_{22}=\alpha\begin{bmatrix}-1 & 0\\  0 & -1 \end{bmatrix}$$
It can be readily verified that $Q$ satisfies the conditions above, i.e,
\begin{itemize}
    \item $Q_{12}Q_{11}^T= \frac{1}{\alpha}\begin{bmatrix} 0 & -1 \\1 & 0\end{bmatrix}$
     \item $ Q_{22}Q_{21}^T= \alpha\begin{bmatrix} 0 & -2 \\2 & 0\end{bmatrix}$
     \item $Q_{22} Q_{11}^T+Q_{21}Q_{12}^T=I$
\end{itemize}
and moreover, the rank of each sub matrix of $Q$ is 2. Denote by
\begin{eqnarray*}
\tilde U &=& UQ_{11}+VQ_{21}\\
\tilde{V} &=& UQ_{12}+VQ_{22}\\
\tilde{U_i} &=& U_iQ_{11}+V_iQ_{21}\\
\tilde{V_i} &=& U_iQ_{12}+V_iQ_{22}
\end{eqnarray*}
Let $i\in [n]$, then $F_i=U_iV^T+V_iU^T=\begin{bmatrix} U_i& V_i\end{bmatrix}\begin{bmatrix} V^T\\ U^T\end{bmatrix}$. As a result $\rank\left(\begin{bmatrix} U_i& V_i\end{bmatrix}\right)\geq \rank(F_i)=2$. Consequently $$ U_i=0\Rightarrow \rank(V_i)=2,V_i=0\Rightarrow \rank(U_i)=2$$ We consider the following 3 possibilities: 
\begin{enumerate}
    \item $U_i=0\Rightarrow \rank(V_i)=\rank(Q_{22})=\rank(Q_{21})=2$. It follows that:
$$\rank(\tilde{U_i})=\rank(U_iQ_{11}+V_iQ_{21})=\rank(V_iQ_{21})=2$$
$$\rank(\tilde{V_i})=\rank(U_iQ_{12}+V_iQ_{22})=\rank(V_iQ_{22})=2$$
    \item $V_i=0\Rightarrow \rank(U_i)=\rank(Q_{11})=\rank(Q_{12})=2$ we get that:
$$\rank(\tilde{U_i})=\rank(U_iQ_{11}+V_iQ_{21})=\rank(U_iQ_{11})=2$$
$$\rank(\tilde{V_i})=\rank(U_iQ_{12}+V_iQ_{22})=\rank(U_iQ_{12})=2$$
\item $\rank(U_i)=\rank(V_i)=2$. Since there exists $j\neq i$ that satisfies one of the two possibilities above it holds that $\rank(\tilde{U_j})=\rank(\tilde{V_j})=2$. This implies, due to Lemma \ref{lemma:rep2}, that $\rank(\tilde{U_i}) \ne 1$ and $\rank(\tilde{V_i}) \ne 1$. Next we exclude the case that either $\tilde U_i=0$ or $\tilde V_i=0$. Assume  w.l.o.g. that $\alpha=1$, $\rank(\tilde{U_i})=2 $ and $\rank(\tilde{V_i})=0$, we have that 
$\tilde{V_i}=0=U_iQ_{12}+V_iQ_{22}$. Now, if we instead select any $\hat\alpha \ne \alpha =1$ then we obtain
$\hat V_i = U_iQ_{12}+ V_i \hat Q_{22}$ where $\hat Q_{22}=\hat\alpha Q_{22}$, from which we obtain $\hat V_i= (\hat\alpha-1) V_iQ_{22} \ne 0$ where the latter inequality is due to the full rank of $V_i$ and $Q_{22}$. 

Next, if $\tilde{U_i}=0$ we can choose yet a different value for $\alpha$, obtaining $\rank(\tilde{U_i})=\rank(\tilde{V_i})=2$. 
Overall, there are at most $2(n-1)$ possible choices of values for $\alpha$ that make either $\tilde{U_i}$ or $\tilde{V_i}$ for some $i \in [n]$ zero. Therefore we can always choose a value for $\alpha$ that will keep all of these blocks rank 2. 
\end{enumerate}
\end{proof}

\section{Supplementary Results}

In the following pages we present additional results for the experiments presented in the main paper. Specifically, we provide tables with complete running time and with median evaluation for the KITTI dataset, in both calibrated and uncalibrated setups (Tables \ref{table:Kitti_Euc}-\ref{table:Kitti_proj_time} in the paper) . We further include here the complete results for the experiments in Table \ref{table::sfm_uncalibrated} in the paper, which involved uncalibrated, unordered internet photos. Below we refer to the  paper "Algebraic Characterization of Essential Matrices and Their Averaging in Multiview Settings" \cite{kasten2019algebraic} as GESFM, which stands for ``Global Essentials SFM".

\begin{table*}[tbh]

\caption{\small  KITTI, calibrated: Mean position error  in meters before BA.  For each dataset and number of cameras we show the median of    mean errors. }\label{table:Kitti_Proj}
\resizebox{\linewidth}{!}{
\begin{tabular}{|c||c|c|c|c|c||c|c|c|c|c||c|c|c|c|c|}\hline
\multirow{2}{2.8em}{\textbf{Dataset}} & \multicolumn{5}{|c||}{ \textbf{5 Cameras}}& \multicolumn{5}{|c||}{ \textbf{10 Cameras}}& \multicolumn{5}{|c|}{\textbf{20 Cameras}} \\\cline{2-16}
& VP &  R4 &  GESFM\cite{kasten2019algebraic} &  LUD\cite{ozyesil2015robust} &1DSFM\cite{wilson2014robust}& VP & R4 & GESFM\cite{kasten2019algebraic} &  LUD\cite{ozyesil2015robust} &1DSFM\cite{wilson2014robust} & VP & R4 &  GESFM\cite{kasten2019algebraic} &  LUD\cite{ozyesil2015robust} &1DSFM\cite{wilson2014robust} \\\hline
 00 
 & \textbf{0.02}

 & \textbf{0.02}
 & {0.30}

 & 0.04 &0.27

 & \textbf{0.04}

 &\textbf{ 0.04}

 & {1.42}
 & 0.06
&0.70
 & \textbf{0.09}

 & \textbf{0.09}

 & {2.92}

 & 0.10
 &1.47
 \\\hline
 01 
 & \textbf{0.37}

 & 1.22
 & {2.47}

 & 1.06
&2.08
 & \textbf{0.63}

 & 3.12

 & {5.70}
 & 1.59
 &4.69
 & \textbf{1.24}

 & 5.82

 & {12.1}

 & 2.32
&4.66
\\\hline
 02 
 & \textbf{0.02}

 & \textbf{0.02}
 & 0.59

 & 0.08
&0.86
 & 0.08

 & \textbf{0.05}

 & {1.90}
 & 0.11
&1.64
 & 0.22

 & \textbf{0.11}

 & {4.46}

 & 0.19
 &1.90
\\\hline
 03 
 & \textbf{0.03}

 & \textbf{0.03}
 & 0.21

 & 0.06
&0.31
 & {0.12}

 & {0.10}

 & 0.52
 & \textbf{0.05}
&0.79
 & {0.42}

 & 1.32

 & 2.33

 & \textbf{0.11}
 &2.11
\\\hline
 04 
 & \textbf{0.03}

 & 0.07
 & 0.83

 & 0.10
&0.79
 & \textbf{0.07}

 & 0.09

 & 2.70
 & 0.14
&1.65
 & \textbf{0.14}

 & 0.76

 & {6.14}

 & 0.28
 &4.94
\\\hline
 05 
 & \textbf{0.01}

 & \textbf{0.01}
 & 0.47

 & 0.04
&0.36
 & \textbf{0.03}

 & \textbf{0.03}

 & {1.51}
 & 0.10
&0.51
 & \textbf{0.08}

 & 0.09

 & {3.33}

 & 0.12
 &1.92
\\\hline
 06 
 & \textbf{0.02}

 & \textbf{0.02}
 & {1.03}

 & 0.09
 &0.80

 & \textbf{0.05}

 & 0.06

 & {2.78}
 & 0.09
&1.48
 & {0.18}

 & 0.76

 & {6.14}

 & \textbf{0.12}
 &3.91
\\\hline
 07 
 & \textbf{0.01}

 & 0.02
 & {0.18}

 & 0.04
&0.21
 & \textbf{0.03}

 & \textbf{0.03}

 & 0.72
 & 0.04
&0.35
 & \textbf{0.08}

 & 0.12

 & {1.53}

 & \textbf{0.08}
 &1.36\\\hline
 08 
 & \textbf{0.02}
 & \textbf{0.02}
 & {0.27}

 & 0.04
&0.61
 & \textbf{0.04}

 & \textbf{0.04}

 & 0.79
 & 0.05
&0.92
 & \textbf{0.09}

 & 0.10

 & {1.87}

 & 0.15
 &2.05
\\\hline
 09 
 & \textbf{0.02}

 & \textbf{0.02}
 & {0.51}

 & 0.09
&0.46
 & {0.05}

 & \textbf{0.04}

 & 2.06
 & 0.07
 &0.60
 & 0.13

 & 0.11

 & {3.85}

 & \textbf{0.10}
 &2.24
\\\hline
 10 
 & \textbf{0.02}

 & \textbf{0.02}
 & 0.42

 & 0.05
& 0.46
 & \textbf{0.03}

 & 0.04

 & {1.64}
 & 0.05
&0.91
 & 0.11

 & 0.12

 & {2.45}

 &  \textbf{0.09} &2.29
\\\hline
\end{tabular}
}
\end{table*}

\begin{table*}[tbh]

\caption{\small KITTI, calibrated: average execution time in seconds, KITTI. \cite{wilson2014robust}'s results are unavailable.}\label{table:Kitti_Proj}
\resizebox{\linewidth}{!}{
\begin{tabular}{|c||c|c|c|c||c|c|c|c||c|c|c|c|}\hline
\multirow{2}{2.8em}{\textbf{Dataset}} & \multicolumn{4}{|c||}{ \textbf{5 Cameras}}& \multicolumn{4}{|c||}{ \textbf{10 Cameras}}& \multicolumn{4}{|c|}{\textbf{20 Cameras}} \\\cline{2-13}
& VP &  R4 &  GESFM\cite{kasten2019algebraic} &  LUD\cite{ozyesil2015robust} & VP & R4 & GESFM\cite{kasten2019algebraic} &  LUD\cite{ozyesil2015robust}  & VP & R4 &   GESFM\cite{kasten2019algebraic} &  LUD\cite{ozyesil2015robust} \\\hline
 00 
 & 0.42
 & 0.48
 & \textbf{0.22}
 & 0.78
 & 1.18
 & 1.18
 & \textbf{0.57}
 & 2.30
 & 2.96
 & 2.24
 & \textbf{1.47}
 & 5.99
 \\\hline
 01 
 & 0.42
 & 0.41
 & \textbf{0.22}
 & 0.19
 & 1.15
 & 0.95
 & \textbf{0.57}
 & 0.77
 & 2.97
 & 1.94
 & \textbf{1.42}
 & 2.29
\\\hline
 02 
 & {0.43}
 & 0.45
 & \textbf{0.22}
 & 0.58
 & 1.16
 & 1.03
 & \textbf{0.57}
 & 1.65
 & 2.97
 & 2.12
 & \textbf{1.45}
 & 4.15
\\\hline
 03 
 & {0.43}
 & 0.55
 & \textbf{0.21}
 & 1.77
 & {1.20}
 & 1.24
 & \textbf{0.57}
 & 5.88
 & {3.20}
 & 2.45
 & \textbf{1.50}
 & 17.65
\\\hline
 04 
 & {0.42}
 & 0.46
 & \textbf{0.22}
 & 0.63
 & {1.22}
 & 1.05
 & \textbf{0.57}
 & 1.94
 & 3.04
 & 2.12
 & \textbf{1.43}
 & 5.29
\\\hline
 05 
 & {0.43}
 & 0.46
 & \textbf{0.22}
 & 0.66
 & 1.21
 & 1.04
 & \textbf{0.62}
 & 2.09
 & 3.15
 & 2.12
 & \textbf{1.49}
 & 5.65
\\\hline
 06 
 & 0.43
 & 0.42
 & \textbf{0.22}
 & 0.33
 & 1.27
 & 0.95
 & \textbf{0.59}
 & 0.95
 & 3.18
 & 1.92
 & \textbf{1.52}
 & 2.54
\\\hline
 07 
 & 0.42
 & 0.49
 & \textbf{0.22}
 & 0.99
 & {1.24}
 & 1.10
 & \textbf{0.60}
 & 3.25
 & 3.07
 & 2.26
 & \textbf{1.48}
 & 9.63\\\hline
 08 
 & {0.42}
 & 0.47
 & \textbf{0.22}
 & 0.79
 & {1.24}
 & 1.07
 & \textbf{0.57}
 & 2.41
 & 3.02
 & 2.16
 & \textbf{1.47}
 & 6.51
\\\hline
 09 
 & {0.41}
 & 0.47
 & \textbf{0.22}
 & 0.71
 & {1.24}
 & 1.07
 & \textbf{0.60}
 & 2.00
 & 3.11
 & 2.11
 & \textbf{1.46}
 & 5.26
\\\hline
 10 
 & {0.43}
 & 0.47
 & \textbf{0.22}
 & 0.84
 & 1.22
 & 1.05
 & \textbf{0.57}
 & 2.53
 & 3.06
 & 2.21
 & \textbf{1.50}
 &  6.68
\\\hline
\end{tabular}
}
\end{table*}

\begin{table*}
\caption{\small KITTI, uncalibrated: Mean reprojection error  in pixels after BA. for each dataset and number of cameras we show the median of    mean errors.}\label{table:Kitti_Proj}
\resizebox{\linewidth}{!}{
\begin{tabular}{|c||c|c|c|c||c|c|c|c||c|c|c|c|}\hline
\multirow{2}{1.em}{\textbf{}} & \multicolumn{4}{|c||}{ \textbf{5 Cameras}}& \multicolumn{4}{|c||}{ \textbf{10 Cameras}}& \multicolumn{4}{|c|}{\textbf{20 Cameras}} \\\cline{2-13}
& VC &  R4 &  GPSFM\cite{kasten2018gpsfm}&   PPSFM\cite{magerand2017practical} & VP & R4 &  GPSFM\cite{kasten2018gpsfm} &   PPSFM\cite{magerand2017practical} & VP & R4 &  GPSFM\cite{kasten2018gpsfm} &   PPSFM\cite{magerand2017practical} \\\hline
 00 
 & \textbf{0.12
}
 & \textbf{0.12}
 & {\textbf{0.12
}}
 & 0.13
 & \textbf{0.15}
 & \textbf{0.15}
 & 0.42 & {0.16
}
 &\textbf{ 0.16
}
 &\textbf{ 0.16}
 & {3.70

}
 & 0.18
 \\\hline
 01 
 &\textbf{0.18}

 & {0.19

} & {0.51
}
 &{0.21}
 & \textbf{0.25}

 & 1.00

 & 1.17
& {0.29
}
 & 5.84
 & 6.77
 & {6.38
}
 &\textbf{ 0.41}
\\\hline
 02 
 & \textbf{0.12
}
 & \textbf{0.12}
 &0.14

 & 0.13
 & \textbf{0.15}
 &\textbf{ 0.15}
 & 2.41
& {0.16
}
 & \textbf{0.16}
 & \textbf{0.16}
 & {10.89

}
 & 0.19
\\\hline
 03 
 & \textbf{0.14
}
 & \textbf{0.14}
 & 0.15

 & 0.16

 & \textbf{0.17
}
 & \textbf{0.17}
 & 0.18 & 0.2
 & \textbf{0.2
}
 & \textbf{0.2}
 & 1.11

 & 0.26
\\\hline
 04 
 & \textbf{0.12
}
 &\textbf{ 0.12}
 & 0.13
 & 0.13
 & \textbf{0.15
}
 & \textbf{0.15}
 & 4.39
 & \textbf{0.15}
 & \textbf{0.18}
 & \textbf{0.18}
 & {27.03

}
 & 0.2
\\\hline
 05 
 & \textbf{0.13
}
 & \textbf{0.13}
 & 0.14
 & 0.15
 &\textbf{ 0.16}
 & \textbf{0.16}
 & 2.74
 & {0.18
}
 &\textbf{ 0.18}
 & \textbf{0.18}
 & {18.73

}
 & 0.22
\\\hline
 06 
 & \textbf{0.12}
 & \textbf{0.12}
 & {0.14
}
 & 0.13
 &\textbf{ 0.15}
 & \textbf{0.15}
 & 0.26

& {0.16
}
 & \textbf{0.17}
 &\textbf{ 0.17}
 & {32.43

}
 & 0.21
\\\hline
 07 
 & \textbf{0.14}

 & \textbf{0.14
}
 & {0.15
}
 & \textbf{0.16}
 & \textbf{0.17
}
 & \textbf{0.17}
& 3.80 & 0.2
 & {0.21}
 & \textbf{0.20}
 & {3.60

}
 & 0.26
\\\hline
 08 
 & \textbf{0.13
}
 & \textbf{0.13
}
 & 0.14
 & \textbf{0.13
} & \textbf{0.16
}
 & \textbf{0.16}
 & 0.29
& 0.18
 & \textbf{0.19}
 &\textbf{ 0.19}
 & {1.85

}
 & 0.21
\\\hline
 09 
 & \textbf{0.11
}
 & \textbf{0.11}
 & {0.13
}
 & 0.12
 & \textbf{0.14
}
 & \textbf{0.14}
 & 1.10
& 0.16
 &\textbf{ 0.16}
 & \textbf{0.16}
 & {7.15

}
 & 0.18
\\\hline
 10 
 & \textbf{0.12
}
 & \textbf{0.12}
 & 0.13
 & 0.13
 & \textbf{0.15}
 & \textbf{0.15}
 & 1.26
& {0.17
}
 &\textbf{ 0.16
}
 &\textbf{ 0.16}
 & {5.65

}
 &  0.19
\\\hline
\end{tabular}
}\end{table*}

\begin{table*}[tbh]

\caption{\small KITTI, uncalibrated: average execution time in seconds. }\label{table:Kitti_Proj}
\resizebox{\linewidth}{!}{
\begin{tabular}{|c||c|c|c|c||c|c|c|c||c|c|c|c|}\hline
\multirow{2}{1.em}{\textbf{}} & \multicolumn{4}{|c||}{ \textbf{5 Cameras}}& \multicolumn{4}{|c||}{ \textbf{10 Cameras}}& \multicolumn{4}{|c|}{\textbf{20 Cameras}} \\\cline{2-13}
& VP &  R4 &  GPSFM\cite{kasten2018gpsfm} &   PPSFM\cite{magerand2017practical} & VP & R4 &  GPSFM\cite{kasten2018gpsfm} &   PPSFM\cite{magerand2017practical} & VP & R4 &  GPSFM\cite{kasten2018gpsfm} &   PPSFM\cite{magerand2017practical} \\\hline
 00 
 & 0.69
 & 1.39
 & \textbf{0.67}
 & 1.99
 & 1.81
 & 2.81
 & \textbf{1.71}
 & 5.50

 & 4.29

 & 5.80

 & \textbf{4.28}

 & 11.05
 \\\hline
 01 
 & 0.56

 & 0.85
 & \textbf{0.50}

 & 2.37

 & 1.48

 & 1.88

 & \textbf{1.31}
 & 5.47
 & 3.89

 & 4.09

 & \textbf{2.82}

 & 10.10
\\\hline
 02 
 & \textbf{0.62}

 & 1.17
 & 0.63

 & 1.71

 & 1.62

 & 2.38

 & \textbf{1.57}
 & 4.55

 & 4.14

 & 5.01

 & \textbf{3.43}

 & 10.00
\\\hline
 03 
 & \textbf{0.83}

 & 1.71
 & 0.95

 & 2.79

 & \textbf{2.08}

 & 3.53

 & 2.33
 & 7.51

 & \textbf{5.17}

 & 8.21

 & 5.84

 & 14.00
\\\hline
 04 
 & \textbf{0.64}

 & 1.18
 & 0.67

 & 1.62

 & \textbf{1.60}

 & 2.35

 & 1.64
 & 4.44

 & 4.21

 & 4.85

 & \textbf{3.27}

 & 10.49
\\\hline
 05 
 & \textbf{0.65}

 & 1.21
 & 0.66

 & 1.84

 & 1.68

 & 2.47

 & \textbf{1.55}
 & 4.67

 & 4.26

 & 5.27

 & \textbf{3.54}

 & 10.64
\\\hline
 06 
 & 0.55

 & 0.93
 & \textbf{0.53}

 & 1.16

 & 1.44

 & 1.92

 & \textbf{1.29}
 & 3.11

 & 3.67

 & 4.15

 & \textbf{2.69}

 & 7.25
\\\hline
 07 
 & 0.75

 & 1.40
 & \textbf{0.70}

 & 3.17

 & \textbf{1.87}

 & 2.97

 & 1.92
 & 11.02

 & 5.04

 & 6.65

 & \textbf{4.52}

 & 22.41
\\\hline
 08 
 & \textbf{0.68}
 & 1.33
 & \textbf{0.68}

 & 1.93

 & \textbf{1.71}

 & 2.75

 & 1.78
 & 5.31

 & 4.57

 & 6.16

 & \textbf{4.41}

 & 10.52
\\\hline
 09 
 & \textbf{0.66}

 & 1.25
 & \textbf{0.66}

 & 2.06

 & \textbf{1.68}

 & 2.54

 & 1.74
 & 5.03
 & 4.33

 & 5.32

 & \textbf{3.91}
 & 11.11
\\\hline
 10 
 & \textbf{0.66}
 & 1.25
 & 0.68
 & 1.80
 & 1.70
 & 2.61
 & \textbf{1.64}
 & 4.94
 & 4.55
 & 5.99
 & \textbf{4.10}
 &  10.42
\\\hline
\end{tabular}
}
\end{table*}

\begin{table*}[tbh]
\tiny
\caption{\small Unordered internet photos, uncalibrated: Mean reprojection
error and execution times.}\label{table:Olsson}
\resizebox{\linewidth}{!}{%
\
\begin{tabular}{|l|r|r|lll|lll|}
\hline
 \multirow{2}{5em}{\textbf{Dataset}} & \multirow{2}{3em} {\textbf{\#points}} &\multirow{2}{3em}{\textbf{\#Images}} &\multicolumn{3}{|c|}{ \textbf{Error(pixels)} }  & \multicolumn{3}{|c|}{\textbf{Time(s)} }  \tabularnewline
\cline{4-9}
 &  &  & \textbf{OURS} & \textbf{GPSFM}\cite{kasten2018gpsfm} & \textbf{\textbf{PPSFM}}\cite{magerand2017practical}  &\textbf{Ours} & \textbf{GPSFM}\cite{kasten2018gpsfm} &\textbf{PPSFM}\cite{magerand2017practical}  \tabularnewline
\hline
Dino 319 & 319 & 36 &  \textbf{0.43} &\textbf{0.43} & 0.47  &  {12.32} &{3.64} & \textbf{3.46}  \tabularnewline
Dino 4983 & 4983 & 36 & {0.43} &\textbf{0.42} & 0.47  &{15.75} &\textbf{4.65} & 13.00   \tabularnewline
Corridor & 737 & 11 & \textbf{0.26} &\textbf{0.26}  & 0.27 & {2.48} &\textbf{1.03} & 1.55  \tabularnewline
House & 672 & 10 & \textbf{0.34} & \textbf{0.34} & 0.40  & 1.82 & \textbf{0.94}& {1.03} \tabularnewline
Gustav Vasa & 4249 & 18 & \textbf{0.16} & \textbf{0.16} & 0.17 &   {4.90} & \textbf{2.47} & 6.64 \tabularnewline
Folke Filbyter & 21150 & 40 & \textbf{0.26} & {0.82} & 0.31 & {14.30} & \textbf{6.70} &  102.77 \tabularnewline
Park Gate & 9099 & 34 & \textbf{0.31} &\textbf{0.31} & 0.45 & {19.68}& \textbf{9.25} & 31.58 \tabularnewline
Nijo & 7348 & 19 & \textbf{0.39} &\textbf{0.39} & 0.44 & {7.02}&\textbf{3.80}& 12.68 \tabularnewline
Drinking Fountain & 5302 & 14 & \textbf{0.28}& \textbf{0.28} & 0.31 & {4.64} & \textbf{2.12} &9.37 \tabularnewline
Golden Statue & 39989 & 18 & \textbf{0.22}& \textbf{0.22} & 0.23 & {10.08} & \textbf{5.05} & 36.21 \tabularnewline
Jonas Ahls & 2021 & 40 & \textbf{0.18}& \textbf{0.18} & 0.20  & {13.84}& \textbf{5.49} & 13.40 \tabularnewline
De Guerre & 13477 & 35 & \textbf{0.26} &\textbf{0.26} & 0.28 & {34.32} &\textbf{11.19}& 32.67 \tabularnewline
Dome & 84792 & 85 & \textbf{0.24} & \textbf{0.24} & 0.25 &{108.18}& \textbf{65.12} & 226.13 \tabularnewline
Alcatraz Courtyard & 23674 & 133 & \textbf{0.52} & \textbf{0.52} & 0.57 & {126.40}& \textbf{63.94} & 151.28 \tabularnewline
Alcatraz Water Tower & 14828 & 172 & \textbf{0.47} & \textbf{0.47} & 0.59 &  169.08 &  90.24 & \textbf{71.80} \tabularnewline
Cherub & 72784 & 65 & {0.75} & \textbf{0.74}& 0.81   &{48.52} &\textbf{27.30}& 101.64 \tabularnewline
Pumpkin & 69335 & 195 & \textbf{0.38} & \textbf{0.38}  &0.44  & {203.06} &\textbf{93.32} & 222.09\tabularnewline
Sphinx & 32668 & 70 & \textbf{0.34}& \textbf{0.34}& 0.36 &  {39.63} & \textbf{31.41} &79.91 \tabularnewline
Toronto University & 7087 & 77 & \textbf{0.24} & 0.54& 0.26  & {30.47}& \textbf{26.59} & 91.26 \tabularnewline

Sri Thendayuthapani & 88849 & 98 & \textbf{0.31} & 0.51 & 0.33   & \textbf{219.11} &{220.25} & 325.58 \tabularnewline
Porta san Donato & 25490 & 141 & \textbf{0.40} & \textbf{0.40} &3.56 
& {126.54} & \textbf{82.18} & 157.96 \tabularnewline
Buddah Tooth & 27920 & 162 & \textbf{0.60} & \textbf{0.60} & 0.71 &  {142.02} & \textbf{59.75} & 81.05 \tabularnewline
Tsar Nikolai I & 37857 & 98 & \textbf{0.29}  & {0.32} & 0.31 
& {89.93} & \textbf{70.79} & 101.01 \tabularnewline
Smolny Cathedral & 51115 & 131 & \textbf{0.46} & {0.48} &  0.50  & {303.62} & \textbf{210.75} & 263.60\tabularnewline
Skansen Kronan & 28371 & 131 & \textbf{0.41} & 0.44 & 0.44  &{118.60} &\textbf{83.43} &161.81 \tabularnewline
\hline
\end{tabular}
}
\end{table*}
\end{appendices}
\end{document}